\documentclass[format=acmsmall,review=false,screen=false]{acmart}

\usepackage[ruled]{algorithm2e}
\usepackage{listings}
\usepackage{graphicx}
\usepackage{program}
\usepackage{ifthen}
\usepackage{epsfig}
\usepackage{array}
\usepackage{color}
\usepackage{hyperref}
\usepackage{url}
\usepackage{xspace}
\usepackage{wasysym}
\usepackage{booktabs}
\usepackage{algorithmic}
\usepackage{textcomp}
\usepackage{comment}
\def\BibTeX{{\rm B\kern-.05em{\sc i\kern-.025em b}\kern-.08em
    T\kern-.1667em\lower.7ex\hbox{E}\kern-.125emX}}

\usepackage{pifont}
\usepackage{fancyvrb}

\newcommand\imcol{\textsc{im2col}\xspace}

\usepackage{subcaption}
\definecolor{blue}{rgb}{0,0,1.0}
\definecolor{darkgreen}{rgb}{0,0.44,0}
\definecolor{green}{rgb}{0,0.44,0}
\definecolor{darkred}{rgb}{0.44,0,0}
\definecolor{darkblue}{rgb}{0,0,0.64}
\definecolor{mygray}{rgb}{0.9,0.9,0.9}
\definecolor{mymauve}{rgb}{0.58,0,0.82}
\definecolor{myred}{rgb}{0.72,0.18,0.0} 
\definecolor{mygreen}{rgb}    {0.0,0.72,0.0} 
\definecolor{myblue}{rgb} {0.18,0.0,0.72} 
\definecolor{mycreme}{rgb}        {1.0,0.8,0.2} 
\definecolor{mygray}{rgb}{0.95,0.95,0.95}
\definecolor{darkgray}{rgb}{0.55,0.55,0.55}

\newcommand{\gemm}{{\sc gemm}\xspace}
\newcommand{\pe}{\mathrel{+\!\!=}}

\graphicspath{{./plots/Results/ARM/}{./plots/Results/Intel/}}

\begin{document}

\title{Algorithm XXX: Automatic Generators for a Family of Matrix Multiplication Routines with Apache TVM}


\author{Guillermo Alaejos}
\affiliation{%
  \institution{Universitat Polit\`ecnica de Val\`encia}
  \city{Valencia}
  \postcode{46022}
  \country{Spain}
}
\email{galalop@upv.es}
\author{Adri\'an Castell\'o}
\affiliation{%
  \institution{Universitat Polit\`ecnica de Val\`encia}
  \city{Valencia}
  \postcode{46022}
  \country{Spain}
}
\email{adcastel@disca.upv.es}
\author{Pedro Alonso-Jord\'a}
\affiliation{%
  \institution{Universitat Polit\`ecnica de Val\`encia}
  \city{Valencia}
  \postcode{46022}
  \country{Spain}
}
\email{palonso@upv.es}
\author{Francisco D. Igual}
\affiliation{%
  \institution{Universidad Complutense de Madrid}
  \city{Madrid}
  \postcode{28040}
  \country{Spain}
}
\email{figual@ucm.es}
\author{H\'ector Mart\'inez}
\affiliation{%
  \institution{Universidad de C\'ordoba}
  \city{C\'ordoba}
  \postcode{14071}
  \country{Spain}
}
\email{el2mapeh@uco.es}
\author{Enrique S. Quintana-Ort\'i}
\affiliation{%
  \institution{Universitat Polit\`ecnica de Val\`encia}
  \city{Valencia}
  \postcode{46022}
  \country{Spain}
}
\email{quintana@disca.upv.es}


\begin{abstract}
We explore the utilization of the Apache TVM open source framework 
to automatically generate a family of algorithms that follow the approach taken by popular linear algebra libraries, such as 
GotoBLAS2, BLIS and OpenBLAS, in order to obtain high-performance blocked 
formulations of the general matrix multiplication (\gemm). 
In addition, we fully automatize the generation
process, by also leveraging the Apache TVM framework to derive a complete variety of the 
processor-specific micro-kernels for \gemm. 
This is in contrast with the convention in 
high performance libraries, which hand-encode 
a single micro-kernel per architecture using Assembly code.
In global, the combination of our TVM-generated blocked algorithms and 
micro-kernels for \gemm 
1)~improves portability, maintainability and, globally, streamlines the software life cycle;
2)~provides high flexibility to easily tailor and optimize the solution to different data types, processor architectures, and matrix operand shapes, yielding 
performance on a par (or even superior for specific matrix shapes) with that of
hand-tuned libraries; and
3)~features a small memory footprint.
\end{abstract}

%
%


 
%
%

\ccsdesc[500]{Mathematics of computing~Mathematical software performance}
\ccsdesc[500]{Computing methodologies~Vector / streaming algorithms}
\ccsdesc[500]{Software and its engineering~Software development techniques}

\keywords{Portability and maintainability, software lifecycle, 
matrix multiplication, BLIS framework, Apache TVM, blocking, SIMD vectorization, high performance}



\maketitle
\renewcommand{\shortauthors}{Guillermo Alaejos et al.}

\section{Introduction}

Over the past decades, there has been a persistent labor toward developing high performance realizations
of linear algebra (LA) libraries for a 
variety of architectures, such as vector processors,
multicore processors with SIMD (single-instruction, multiple-data) units,
data-parallel graphics processing units (GPUs) and, more recently, 
domain-specific architectures and accelerators.
This effort has come from major hardware vendors, with some relevant products being Intel MKL, 
AMD AOCL, IBM ESSL, ARMPL and NVIDIA cuBLAS, as well as from the academic side, with software packages such
as GotoBLAS2~\cite{Goto:2008:AHP}, 
OpenBLAS~\cite{OpenBLAS:ICPADS}, 
BLIS~\cite{BLIS1} and
ATLAS~\cite{ATLAS}. 

The general
matrix multiplication (\gemm) is a crucial computational kernel upon which these LA libraries
are built. In addition, \gemm is also a key operation for deep learning applications that leverage 
transformers for natural language processing or convolutional deep neural networks 
for signal processing and computer vision~\cite{8114708,DBLP:journals/corr/abs-1802-09941}.
Unfortunately, 
these LA libraries present a few obstacles:
\begin{enumerate}
\item The optimized routines in these libraries are hardware-specific. 
      This is the case for Intel, IBM, ARM 
      and NVIDIA's products. To a lesser
      extent, it also applies to GotoBLAS2, OpenBLAS and BLIS,\footnote{The same comment applies to AMD's library, which is simply a customized version of BLIS.} which leverage a collection of processor-customized micro-kernels~\cite{BLIS2}.
\item Developing highly optimized micro-kernels for \gemm requires
      a deep knowledge of high performance computing and computer architecture.
\item The code in these libraries is very large. Besides, it is 
      hard to master due
      to the abundant use of productivity-enhancing 
      macros, templates and 
      high level programming techniques.
      Maintaining the libraries thus mostly
      lies in the hands of the original developers.
\item The software {misses some relevant 
      cases} such as, for example, support for half (i.e., 16-bit) floating point precision or integer arithmetic.
\item The implementation of \gemm in these libraries is sub-optimal under certain circumstances as 
      the code is 
      usually tuned for ``squarish'', large-scale cases. 
\item The memory footprint of the libraries is often 
      in the order of Mbytes.  
\end{enumerate}

In this paper we address the limitations of LA libraries by demonstrating that it is possible to automatically generate a family of blocked algorithms for \gemm, together with a collection of micro-kernels for \gemm, using Apache TVM \textcolor{black}{(Tensor Virtual Machine)}~\cite{TVM_1}.
This alternative solution offers
the following advantages:
\begin{enumerate}
 \item At a high level, the library is ``replaced'' by a collection of TVM generators (in the form of Python routines), reducing the maintainability effort to a bare minimum
       and largely improving the portability of the solution.
 \item Using the appropriate backend compiler, 
       the generation/optimization can be 
       easily specialized for distinct data types, 
       further enhancing the portability and maintainability of the solution.
 \item By adjusting the algorithm and micro-kernel 
       to the problem, it is possible
       to outperform high performance realizations of \gemm in 
       commercial products as well as academic libraries.
 \item 
       The optimization process for each problem dimension is largely seamless, boiling down to the evaluation of 
        a reduced number of micro-kernels.
        In other words, the optimization search space is limited.
 \item The memory footprint of the resulting realization of \gemm is very small and the entire framework library is also very small.
\end{enumerate}

\textcolor{black}{Our paper makes two significant contributions. Firstly, it offers a comprehensive and pedagogical tutorial on building a high-performance implementation of \gemm using TVM, complete with extensive code examples and detailed explanations. Secondly, and perhaps more importantly, it provides compelling evidence for the advantages of automating the development process. These advantages include enhanced portability, reduced maintenance costs, insulation from hardware-specific optimization details, and ultimately, improved performance. This performance boost arises from the ability to explore multiple micro-kernels, which outperforms libraries relying on a single general-purpose micro-kernel.}

The rest of the paper is structured as follows.
Section~\ref{sec:related_work} provides an in-depth review of 
the state-of-the-art in automatic realizations of high performance
libraries in the fields of dense linear algebra in general, and 
deep learning in particular, and discusses different alternatives towards 
automatic code generation.
In Section~\ref{sec:family} and~\ref{sec:microkernels}
we respectively review the modern realization of algorithms 
and micro-kernels for \gemm.
This is then followed, 
in Sections~\ref{sec:tvm-family} and~\ref{sec:tvm-microkernels}, 
by the introduction of the automatic generation of a baseline \gemm algorithm and a variety of micro-kernels, respectively;
and in 
Section~\ref{sec:otherfamily}, by the extension of 
the automatization techniques to cover a complete family of \gemm algorithms.
In Section~\ref{sec:experiments} we evaluate the performance of the resulting solution in 
a platform equipped with ARM cores, and demonstrate its portability to other modern architectures
from different vendors. 
Finally, in Section~\ref{sec:remarks} we summarize the main results from this work. 

\section{Related work\label{sec:related_work}}

Automatic code generation
is gaining interest in the last years as a means 
to attain performance portability across existing and new 
architectures, for deep learning (DL) models, with a minimal 
intervention from the programmer~\cite{DBLP:journals/corr/abs-2002-03794}. 
Recent languages and compiler frameworks, such as Halide~\cite{Halide} 
or TVM~\cite{TVM_1}, propose a clear separation of concerns between 
the definition of the operation and its optimization, in order to ease 
development of operators to a plethora of target 
architectures, including general-purpose processors, GPUs, digital signal processors (DSPs), 
and specific-purpose accelerators~\cite{MoreauVTA2018}.
Starting from a {\em computational graph} which defines the operator and
the {\em data flow}, optimization techniques for 
performance portability are applied 
at the graph level, via operator fusions and transformations; 
and at the operator level, with hardware-specific optimizations.
Some of these optimizations are framework-specific
(e.g., TensorFlow XLA \cite{XLA}) while others, developed by experts, are realized
within specialized libraries (e.g., NVIDIA cuDNN\cite{cuDNN}).

From a technical perspective, DL compilers can be broadly classified
as JIT ({\em Just-in-time}) or AOT ({\em Ahead-of-time}).
JIT compilers generate executable code on-the-fly. Hence, they can exploit 
extended runtime knowledge to fine-tune the final executable, at the
cost of a certain overhead due to the code generation
logic. Many common DL frameworks and compilers rely or support JIT, including
TFLite and its Micro variant\footnote{http://www.tensorflow.org/lite.}, XLA, MLIR \cite{MLIR} and TVM.
Contrarily, AOT compilers generate executable code a priori, and execute common 
versions at runtime without further modification. The advantage of the AOT approach is 
two-fold: first, it can extend the analysis scope, hence accommodating more 
thorough optimizations; second, it eases the development of cross-compilation 
schemes for embedded architectures \cite{C-GOOD}
or remote execution-only architectures \cite{TVM_2}.
The use of external libraries for DL primitives 
(mainly convolutions and linear algebra primitives) can be also considered as AOT, 
since in general they are implemented statically and optimized prior to execution time.

Armed with hardware-agnostic IRs (Intermediate Representations), 
these compiler frameworks effectively decouple schedule from computation, 
and enable the automatic exploration of the scheduling and configuration
spaces by means of auto-tuning techniques. A clear example is 
AutoTVM~\cite{Chen18_Learning}, integrated within TVM, which performs
a complete exploration of the search space; at each iteration, 
the framework tests the generated code on the target device, and stores some feedback
which is eventually leveraged to find the optimal combination of configuration parameters.
This {\em brute force} optimization scheme guarantees finding the optimal
solution (configuration setup), but the number of tested configurations
grows exponentially with the dimension of the design space.
Hence, the use of these naive schemes is limited to problems with reduced
search spaces, or where online testing is time-inexpensive.
(Unfortunately, this is not the case of 
the architecture-aware adaptation of DL models to a specific 
hardware setup.)
In order to alleviate the expensive search-and-test procedure, automatic 
learning schemes (such as XGBoost \cite{ChenG16_XGBoost}, within TVM) have been 
enhanced \cite{Zhang2019_GEMM} with
Random Search \cite{Bergstra2012}, Bayesian optimization \cite{Bayesian_1}
and Genetic algorithms \cite{Genetic_1}. 
More recently, Deep Reinforcement Learning techniques have been proposed
to autonomously guide and extract policies for optimal execution 
configuration \cite{RL1,RL2,RL3,RL4}.
While all these efforts alleviate the cost of the 
hyperparameter search, they are still computationally expensive
and, in many cases, yield solutions that are difficult to explain or
reproduce for developers, or to port to embedded architectures or systems
with reduced compute capabilities.

Auto-tuning techniques within DL compiler frameworks follow
similar ideas to those of auto-tuning in dense LA routines.
ATLAS \cite{ATLAS} selects the cache configuration parameters and 
the micro-kernel at installation time via a
comprehensive search procedure. 
Recently, in the framework of the BLIS project, the work in \cite{BLIS4} demonstrated that 
deriving analytical models for optimal 
configuration parameters selection is an effective way to attain high performance
without the necessity of an auto-tuning scheme for configuration parameter exploration.
Replacing the auto-tuning scheme with model-based solutions
was also successfully explored in~\cite{OpenBLAS:ICPADS,Yotov:2005}.

Our approach combines the advantages of existing JIT compiler frameworks for easily
deriving high performance codes for \gemm-based DL primitives, with 
analytical models 
in order to avoid expensive auto-tuning.
Our work differs from \cite{GEMM_MLIR}, which uses MLIR to 
describe early experiences exclusively with \gemm,
as well as from \cite{Zhang2009}, which proposes advanced
auto-tuning schemes for this primitive.
Concretely, we leverage TVM to extend and
further analyze the automatic generation of \gemm-based
primitives for DL, integrating analytical models to ease
the optimization process without the need of expensive auto-tuning
procedures. In addition, we apply our ideas to 
\gemm (Section~\ref{sec:tvm-family} and~\ref{sec:tvm-microkernels}). 
%

The use of analytical models to determine optimal execution
parameters eliminates the time penalty introduced by
auto-tuning, and avoids the lack of accuracy and reproducibility.
%
While not discussed in this work, 1) extending the analytical models to address other algorithmic variants for \gemm,
which favor certain cache hierarchy setups \cite{CasDQ22},
as additional configuration 
parameters; and 2) supporting optimal parallelization
schemes \cite{BLIS3} for \gemm-based operators,
and supporting this additional degrees of freedom
within the compilation stage are also clear 
benefits of our proposal. In comparison, these extensions would introduce a non-affordable
complexity to auto-tuning schemes.

\section{A Family of Blocked Algorithms for \gemm}
\label{sec:family}

In this section we review the modern realization of 
\gemm in current high performance libraries, and generalize
the ideas to discuss 
a full family of algorithms for this operation.

\subsection{The baseline algorithm}

Consider the 
matrix multiplication $C = C + AB$, 
abbreviated as $C \pe AB$,
where the matrix operands present the following dimensions:
$A \rightarrow m \times k$, 
$B \rightarrow k \times n$, and 
$C \rightarrow m \times n$.
The realization of this kernel for general-purpose 
processor architectures
with hierarchically-layered memories,
in libraries such as OpenBLAS, BLIS, 
AMD AOML and, possibly, Intel MKL/oneMKL, 
follow the basic ideas of GotoBLAS2~\cite{Goto:2008:AHP} to
decompose the computation into five nested loops, traversing the $m, n, k$ dimensions of the problem in a specific order. 
Inside these loops, two packing routines copy (and re-arrange) certain blocks of the input operands $A,~B$ 
into two buffers, 
$A_c \rightarrow m_c \times k_c$, $B_c \rightarrow k_c \times n_c$, 
in order to favor an efficient utilization
of the cache memories. 
(For simplicity, 
in the following we assume that $m$, $n$, and $k$ are integer multiples of $m_c$, $n_c$, and $k_c$,
respectively.) 
Furthermore, the fifth loop comprises a micro-kernel that is often vectorized 
to exploit the SIMD units in most current general-purpose processors~\cite{BLIS1}. 
In short detail, 
the \textit{baseline algorithm for \gemm} 
aims at maintaining a block $B_c$ in the L3 cache and a block $A_c$ in the L2 cache,
while streaming $C$ directly from the main
memory into the processor registers. Furthermore, a small
micro-panel of 
$B_c$ is to reside in the L1 cache.

\begin{figure*}[tbp!]
\centering
\begin{tabular}{c}
\includegraphics[width=0.9\textwidth]{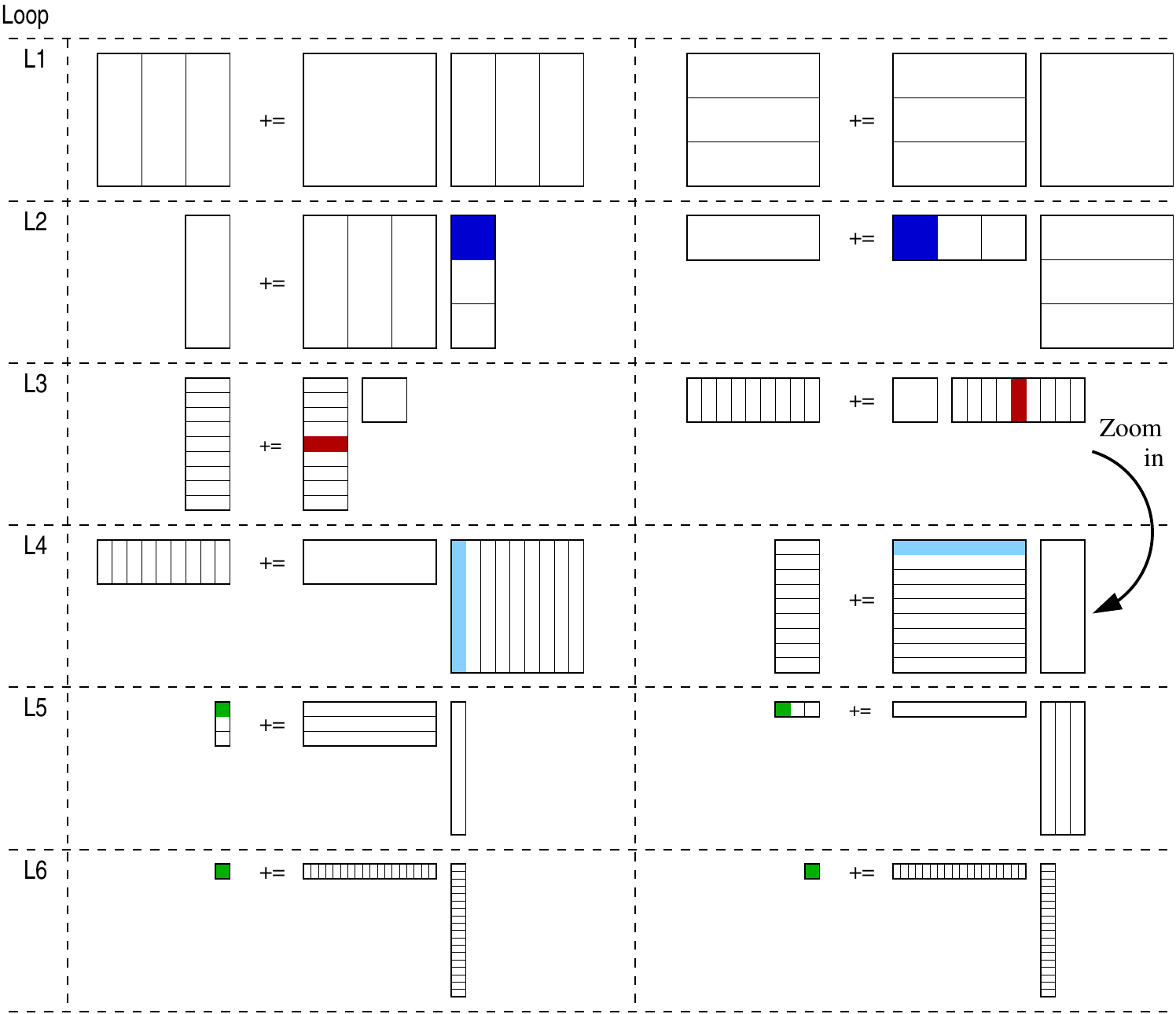}\hspace*{0.8cm} \\
~\\
\includegraphics[width=0.7\textwidth]{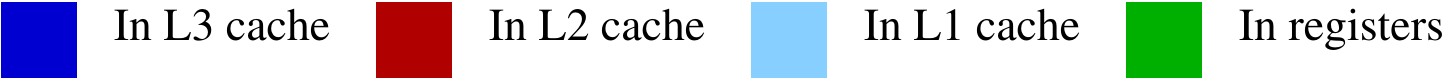}\hspace{0.8cm}
~\\
~\\
~\\
\begin{tabular}{ccc}
\begin{minipage}[c]{0.46\textwidth}
\footnotesize
\begin{lstlisting}[language=C,alsoletter={.},deletekeywords={.sum},morekeywords={}]
for (jc=0; jc<n; jc+=nc) // Loop L1
 for (pc=0; pc<k; pc+=kc) {   // L2
  // Pack B
  Bc := B(pc:pc+kc-1,jc:jc+nc-1);
  for (ic=0; ic<m; ic+=mc) {  // L3
   // Pack A
   Ac := A(ic:ic+mc-1,pc:pc+kc-1);
   for (jr=0; jr<nc; jr+=nr)  // L4
    for (ir=0; ir<mc; ir+=mr) // L5
     // Micro-kernel
     C(ic+ir:ic+ir+mr-1, 
       jc+jr:jc+jr+nr-1)
       += Ac(ir:ir+mr-1,0:kc-1)
       *  Bc(0:kc-1,jr:jr+nr-1);
 }}
\end{lstlisting}
\end{minipage}
&~&
\begin{minipage}[c]{0.46\textwidth}
\footnotesize
\begin{lstlisting}[language=C,alsoletter={.},deletekeywords={.sum},morekeywords={}]
for (ic=0; ic<m; ic+=mc) // Loop L1
 for (pc=0; pc<k; pc+=kc) {   // L2
  // Pack A
  Ac := A(ic:ic+mc-1,pc:pc+kc-1);
  for (jc=0; jc<n; jc+=nc) {  // L3
   // Pack B
   Bc := B(pc:pc+kc-1,jc:jc+nc-1);
   for (ir=0; ir<mc; ir+=mr)  // L4
    for (jr=0; jr<nc; jr+=nr) // L5
     // Micro-kernel
     C(ic+ir:ic+ir+mr-1, 
       jc+jr:jc+jr+nr-1)
       += Ac(ir:ir+mr-1,0:kc-1)
       *  Bc(0:kc-1,jr:jr+nr-1);
 }}
\end{lstlisting}
\end{minipage}
\end{tabular}
\end{tabular}
\caption{Baseline (B3A2C0) and A3B2C0 algorithms (left and right, respectively) for \gemm.}
\label{fig:blis_family_Cresident}
\end{figure*}

The orchestration of the data movements across the memory
hierarchy in the baseline algorithm for \gemm is endowed by the specific nesting of the algorithm loops, 
in combination with a careful choice of the
loop strides/sizes of the buffers~\cite{BLIS4}. 
The operands' partitioning induced by the loops as well as the target level of the memory hierarchy for each operand block
are illustrated in Figure~\ref{fig:blis_family_Cresident} (left).
We will also refer to the baseline algorithm as B3A2C0, 
where each letter denotes one of the three matrix 
operands, and the subsequent number specifies the cache level where 
a part of the operand is to reside, 
with~0 referring to the processor registers. As $B$ is to reside both in the L1 and L3 caches, we do not
specify the former in the notation.

\subsection{Other members of the \gemm family}

\begin{figure}[tbp!]
\centering
\begin{tabular}{c}
\includegraphics[width=0.9\textwidth]{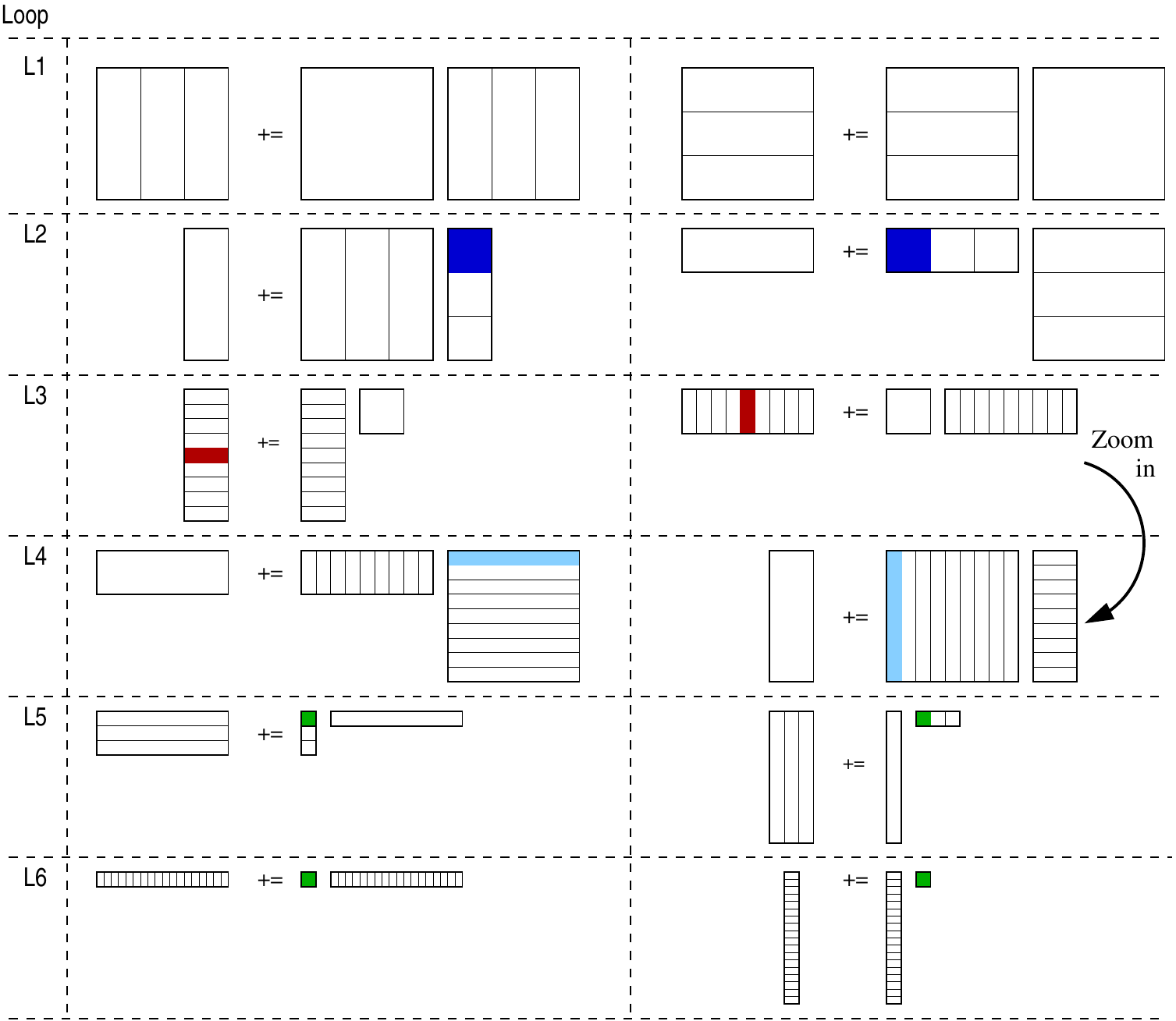}\hspace*{0.8cm} \\
~\\
\includegraphics[width=0.7\textwidth]{Figures/blis_family_legend2.pdf}\hspace{0.8cm}
~\\
~\\
~\\
\begin{tabular}{ccc}
\begin{minipage}[c]{0.46\textwidth}
\footnotesize
\begin{lstlisting}[language=C,alsoletter={.},deletekeywords={.sum},morekeywords={}]
for (jc=0; jc<n; jc+=nc) // Loop L1
 for (pc=0; pc<k; pc+=kc) {   // L2
  // Pack B
  Bc := B(pc:pc+kc-1,jc:jc+nc-1);
  for (ic=0; ic<m; ic+=mc) {  // L3
   // Pack C
   Cc := C(ic:ic+mc-1,jc:jc+nc-1);
   for (pr=0; pr<kc; pr+=kr)  // L4
    for (ir=0; ir<mc; ir+=mr) // L5
     // Micro-kernel
     Cc(ir:ir+mr-1,0:nc-1)
        += A(ic+ir:ic+ir+mr-1,
             pc+pr:pc+pr+kr-1)
        *  Bc(pr:pr+kr-1,0:nc-1);
   // Unpack C
   C(ic:ic+mc-1,jc:jc+nc-1) := Cc;
 }}
\end{lstlisting}
\end{minipage}
&~&
\begin{minipage}[c]{0.46\textwidth}
\footnotesize
\begin{lstlisting}[language=C,alsoletter={.},deletekeywords={.sum},morekeywords={}]
for (ic=0; ic<m; ic+=mc) // Loop L1
 for (pc=0; pc<k; pc+=kc) {   // L2
  // Pack A
  Ac := A(ic:ic+mc-1,pc:pc+kc-1);
  for (jc=0; jc<n; jc+=nc) {  // L3
   // Pack C
   Cc := C(ic:ic+mc-1,jc:jc+nc-1);
   for (pr=0; pr<kc; pr+=kr)  // L4
    for (jr=0; jr<nc; jr+=nr) // L5
     // Micro-kernel               
     Cc(0:mr-1,jr:jr+nr-1) 
        += Ac(0:mr-1,pr:pr+kr-1) 
        *  B(pc+pr:pc+pr+kr-1,
             jc+jr:jc+jr+nr-1);
   // Unpack C
   C(ic:ic+mc-1,jc:jc+nc-1) := Cc;
 }}   
\end{lstlisting}
\end{minipage}
\end{tabular}
\end{tabular}
\caption{B3C2A0 and A3C2B0 algorithms (left and right, respectively) for \gemm.}
\label{fig:blis_family_CL2}
\end{figure}

We continue the discussion 
on the high performance realization of \gemm by noting that
there exist five other algorithmic variants 
which can be obtained by re-organizing the loops of the baseline algorithm in a different manner~\cite{DBLP:journals/corr/abs-1904-05717,10.1007/11558958_30,CasDQ22}.
For example, a ``twin'' version 
is directly obtained by swapping the roles of $A$ and $B$ in the 
baseline algorithm, 
yielding the A3B2C0 variant,
where two blocks of $A$ respectively occupy the L1 and L3 caches and a block of $B$ resides in the L2 cache;
see Figure~\ref{fig:blis_family_Cresident} (right).
In the same line, 
Figure~\ref{fig:blis_family_CL2} 
displays two additional variants of the \gemm family: 
B3C2A0 (left) and A3C2B0 (right) that maintain a block of $C$ in the L2 cache.
The two missing variants, where $C$ resides in the L3 cache, 
(C3B2A0 and C3A2B0, omitted for brevity,) are derived 
by swapping the roles of $A/B$ with $C$ in the two algorithms 
given in that figure; 
see~\cite{CasDQ22}.

\newcommand{\Aresident}{Resident~$A$\xspace}%
\newcommand{\Bresident}{Resident~$B$\xspace}%
\newcommand{\Cresident}{Resident~$C$\xspace}%
\newcommand{\ABresident}{Resident~$A/B$\xspace}%

\section{High performance micro-kernels for \gemm}
\label{sec:microkernels}

In this section, we connect the six blocked algorithms 
for \gemm with three types of micro-kernels
that differ in the matrix operand that resides in the processor
registers~\cite{DBLP:journals/corr/abs-1904-05717,10.1007/11558958_30,CasDQ22}. For simplicity, we assume that 
the cache configuration parameters
$m_c$,
$n_c$,
$k_c$ are respectively integer multiples of the 
micro-kernel parameters
$m_r$,
$n_r$,
$k_r$
where, depending on the type of micro-kernel, two of the latter parameters
specify the dimension of the micro-tile that resides in
the processor registers.

\subsection{Operand \Cresident (in the processor registers)}

The baseline algorithm for \gemm
and its ``twin'' A3B2C0 (both
in Figure~\ref{fig:blis_family_Cresident})
cast the innermost computation in terms
of a micro-kernel that computes 
a smaller \gemm, $C_r \pe A_rB_r$, 
where 
$A_r \rightarrow m_r \times k_c$,
$B_r \rightarrow k_c \times n_r$ respectively denote two micro-panels of the buffers $A_c,B_c$; while
$C_r \rightarrow m_r \times n_r$ is a small micro-tile of $C$ that resides in the processor registers during the execution of the micro-kernel.
This corresponds to the operation performed inside loop \texttt{L5} 
of the baseline algorithm B3A2C0
(see Figure~\ref{fig:blis_family_Cresident}),
with 
\begin{list}{}{}
\item $A_r =$~\texttt{\small Ac(ir:ir+mr-1,0:kc-1)},
\item $B_r =$~\texttt{\small Bc(0:kc-1,jr:jr+nr-1)},~~ and
\item $C_r =$~\texttt{\small C(ic+ir:ic+ir+mr-1,jc+jr:jc+jr+nr-1)}.
\end{list}
The realization of this micro-kernel 
iterates across the $k_c$ dimension of the problem
(as part of an additional loop, labeled as \texttt{L6}), at each step performing an outer product 
involving a single column of $A_r$ and a single row of $B_r$ to update the entire micro-tile $C_r$; see Figure~\ref{fig:blis_microkernel_CABresident} (top).


For high performance, the data in $A_c,B_c$ are carefully packed
as illustrated in Figure~\ref{fig:blis_packing}, in order
to ensure access with 
unit stride to the columns of $A_r$ and the rows of $B_r$ from
within the micro-kernel. 
This reduces the number of cache evictions during these accesses as well as 
accommodates the use of efficient SIMD instructions to load these elements into vector registers and operate with them.

\subsection{Other types of micro-kernels: Operand \Aresident or \Bresident}

\begin{figure}[tbp!]
\centering
\begin{tabular}{ccc}
\begin{minipage}[c]{0.45\textwidth}
\footnotesize
\begin{lstlisting}[language=C,alsoletter={.},deletekeywords={.sum},morekeywords={}]
for (pr=0; pr<kc; pr++) // Loop L6
  C(ic+ir:ic+ir+mr-1,
    jc+jr:jc+jr+nr-1)
    += Ac(ir:ir+mr-1,pr)
    *  Bc(pr,jr:jr+nr-1);
\end{lstlisting}
\end{minipage}
&~~~&
\includegraphics[width=0.30\textwidth]{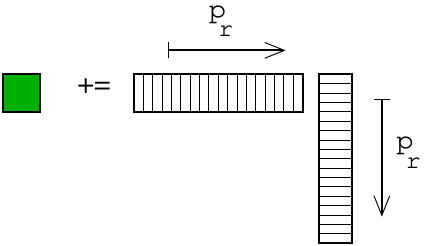}
\\
\\\hline
\\
\begin{minipage}[b]{0.45\textwidth}
\footnotesize
\begin{lstlisting}[language=C,alsoletter={.},deletekeywords={.sum},morekeywords={}]
for (jr=0; jr<nc; jr++) // Loop L6
  Cc(ir:ir+mr-1,jr)
     += A(ic+ir:ic+ir+mr-1,
          pc+pr:pc+pr+kr-1)
     *  Bc(pr:pr+kr-1,jr);
\end{lstlisting}
\end{minipage}
&~~~&
\includegraphics[width=0.40\textwidth]{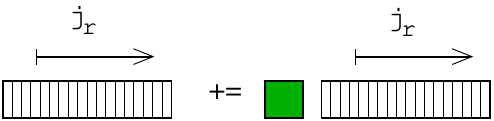}
\\
\\\hline
\\
\begin{minipage}[b]{0.45\textwidth}
\footnotesize
\begin{lstlisting}[language=C,alsoletter={.},deletekeywords={.sum},morekeywords={}]
for (ir=0; ir<mr; ir++) // Loop L6
  Cc(ir,jr:jr+nr-1)
     += Ac(ir,pr:pr+kr-1)
     *  B(pc+pr:pc+pr+kr-1,
          jc+jr:jc+jr+nr-1);
\end{lstlisting}
\end{minipage}
&~~~&
\includegraphics[width=0.32\textwidth]{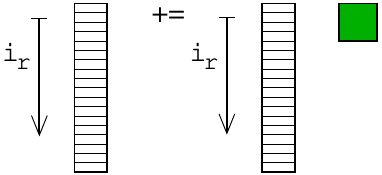}
\end{tabular}
\caption{Micro-kernels with \Cresident, \Aresident or \Bresident 
         in the processor registers (top, middle and bottom, respectively).}
\label{fig:blis_microkernel_CABresident}
\end{figure}

The four remaining variants of \gemm 
(see, e.g., Figure~\ref{fig:blis_family_CL2})
leverage two other types
of micro-kernels: 
The first one 
maintains a micro-tile of $A$ in the processor registers while
performing
an $m_r \times k_r$ matrix-vector product per iteration of loop~\texttt{L6}. 
The second one 
keeps 
a micro-tile of \Bresident in the processor registers, and
carries out
an $k_r \times n_r$ vector-matrix product per iteration of loop~\texttt{L6}. 
These two types of micro-kernels are illustrated in
Figure~\ref{fig:blis_microkernel_CABresident} (middle and
bottom).

\begin{figure}[tbp!]
\centering
\begin{tabular}{c}
\includegraphics[width=0.5\textwidth]{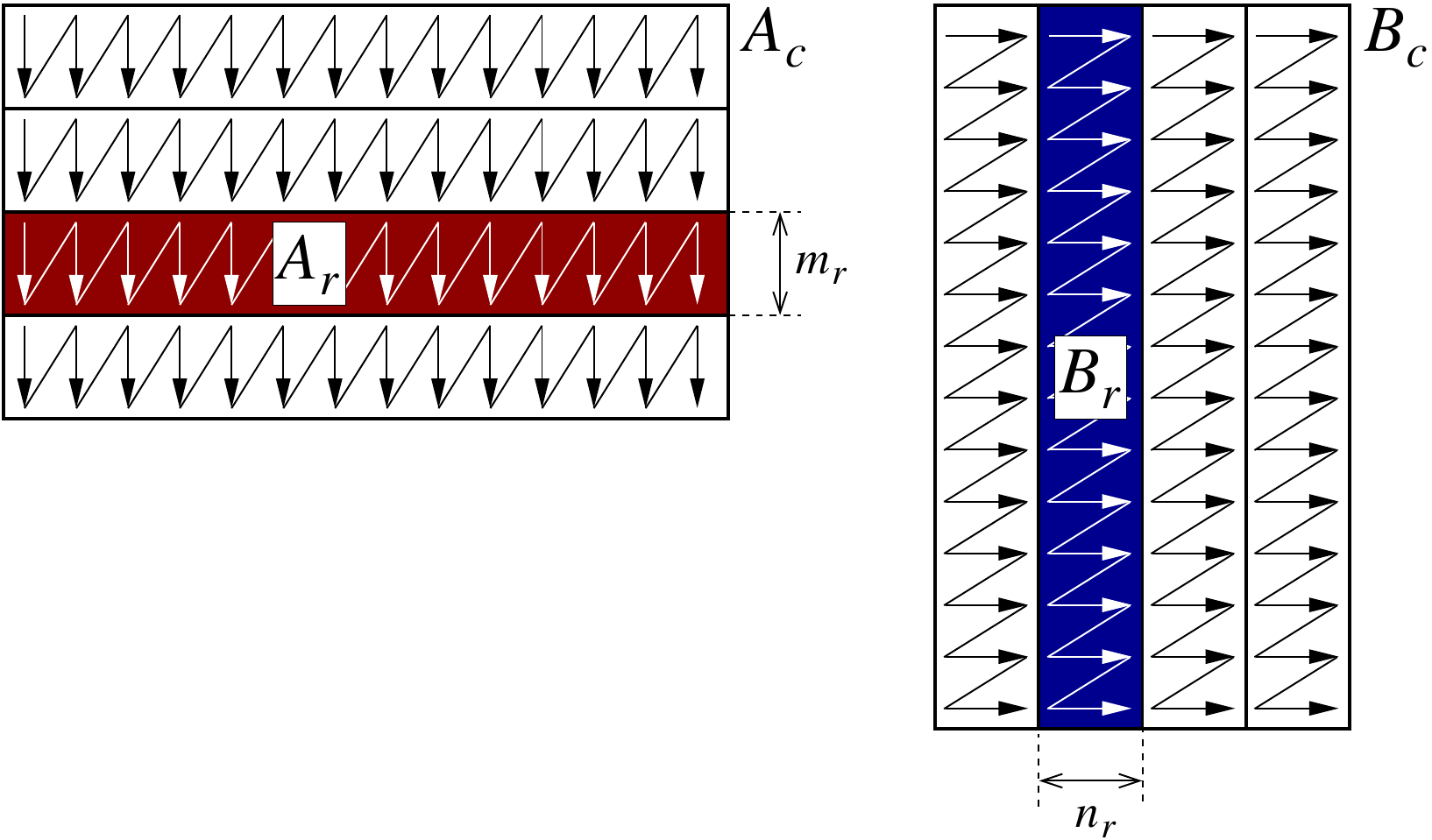}
\end{tabular}
\caption{Packing in the baseline algorithm.}
\label{fig:blis_packing}
\end{figure}

To enable SIMD loads/stores, 
each type of micro-kernel requires a specialized packing scheme for two 
of the matrix operands.
For the micro-kernel with \Aresident (in the processor registers),
$C_c$ and $B_c$ are packed following
the same pattern as $A_c$
in Figure~\ref{fig:blis_packing} (with the entries of $B_c$ arranged into micro-panels of $k_r$ rows). For the micro-kernel with \Bresident, the
buffers for $C_c$ and $A_c$ are packed as $B_c$ 
in the same figure
(with the entries of $A_c$ arranged into micro-panels of $k_r$ columns).


\subsection{High performance}

A few rules of thumb guide the design of a high performance micro-kernel for a processor architecture with a multi-layered memory hierarchy~\cite{BLIS4}. We discuss them for a micro-kernel with \Cresident, 
but they are easily  to derive for the two other types of micro-kernel:
 \begin{itemize}
\item Considering the $k_c$ successive updates of the micro-tile $C_r$ occurring in loop~\texttt{L6},
      the micro-kernel parameters $m_r,n_r$ should be chosen sufficiently large so as to avoid stalls due
      the latency between the issuance of two instructions that update the same entry of $C_r$.
\item Ideally, $m_r$ should be equal to $n_r$ as this maximizes the ratio of computation to 
      data movement during the update of $C_r$ in loop~\texttt{L6}.
\end{itemize}
These two principles suggest maximizing the values for $m_r,n_r$ as part of a ``large'' micro-kernel. 
In practice though, 
the limited number of vector registers 
constrains the practical values of $m_r,n_r$ 
within a couple of dozens. 

A comparison of the micro-kernel
with \Cresident and those with \ABresident reveals some
differences:
\begin{itemize}
\item The variants with \Cresident presents a higher
      arithmetic intensity since, while all types of micro-kernels
      perform the same number of flops and number of loads 
      from memory, the variants with \ABresident have 
      to write a column or row of $C$ back into the memory at each iteration of the loop.
\item \textcolor{black}{Unlike the micro-kernel with \Cresident, the variants with \ABresident 
      do not present a RAW (read-after-write) dependency between consecutive
      iterations of the micro-kernel.} 
\end{itemize}

The implementation of the micro-kernels in general-purpose
processor architectures equipped with SIMD
arithmetic units is in practice done in Assembly code; 
vectorized using architecture-specific SIMD instructions (e.g., Intel SSE/AVX, ARM NEON, etc.); 
and  enhanced with high performance
computing techniques such as loop unrolling, software pipelining, data prefetching, etc.~\cite{Dowd98}.

We close this section by noting that for large, squarish \gemm problems, the optimal values
for the micro-kernel parameters $m_r,n_r,k_r$ can be determined, for a given
processor architecture, via a few experimental tests. However, \textcolor{black}{for problems with one dimension much
larger than the others}, the optimal values of these parameters may vary considerably, and they can
only be determined experimentally by first implementing them. Unfortunately, developing a 
high performance micro-kernel is mostly a manual process, which requires an expert
with a deep knowledge of high performance computing and computer architecture
to attain optimal performance.

\section{Automatic Generation of the Baseline Algorithm for \gemm}
\label{sec:tvm-family}

Apache TVM is an open source compiler framework that allows generating, optimizing, and executing 
machine learning routines for (general-purpose) 
multicore processors, GPUs (graphics processing units), and other accelerator 
backends~\cite{10.5555/3291168.3291211}.
In our effort toward automatically generating high performance algorithms for \gemm, 
we start from the TVM tutorial 
on how to optimize \gemm on a 
general-purpose processor by blocking (tiling) and
packing the matrix operands.\footnote{\url{https://tvm.apache.org/docs/tutorials/}.}
Concretely, we 
build upon these instructions to assemble a TVM generator 
that automatically produces the baseline algorithm for \gemm. 
{\color{black} Our approach extends the tutorial in that we apply the optimizations in a BLIS-like manner instead of following a general optimization approach. More concretely we introduce the cache-aware packings as well as software prefetching. In addition, we provide guidelines to derive different algorithmic variants for \gemm.}

\subsection{Basic \gemm with TVM}

Consider a basic realization of \gemm consisting of three loops that compute each element 
of the output matrix
by performing a reduction (i.e., an inner or dot product) across the $k$ dimension of the problem:
\vspace*{-2ex}
\begin{center}
\begin{minipage}[t]{0.6\textwidth}
\begin{lstlisting}[language=C,alsoletter={.},deletekeywords={.sum},morekeywords={},backgroundcolor=\color{white},numbers=none,frame=none]
for (i=0; i<m; i++) 
  for (j=0; j<n; j++) 
     for (p=0; p<k; p++) 
       C(i,j) += A(i,p) * B(p,j);
\end{lstlisting}
\end{minipage}
\end{center}
This specific realization of \gemm, with the loops traversing the dimensions of the problem in the order 
$i\Rightarrow j\Rightarrow p$, can be 
obtained using the (Python-based) TVM generator in
Figure~\ref{lst:basic_GEMM}. We distinguish 
\textcolor{black}{nine} fundamental parts in the code (in this basic example, some of the parts
are empty, but they will appear
in the refined versions of the generator described in subsequent sections):
\begin{description}
\item[\em Part \textsf{P0}. Parameter list] The function receives the matrix dimensions {\tt m, n, k}, and the data type of the operands ({\tt dtype})
as parameters.
\item[\em Part \textsf{P1}. Declaration of the input operands:] Lines 3 and 4 define two 
operands, or placeholders, respectively for $A$ and $B$, 
of the appropriate dimensions and  data type.
\item[\em Part \textsf{P2}. Definition of the operation:]
      Lines 7--10 specify the operation to be computed in terms of the 
      two placeholders. 
      In particular, line~7 defines the computation in terms of a reduction ({\tt sum}) across the dimension {\tt k}.
       
\item[\em Part \textsf{P3}. Preparation of the schedule:] 
      Line 13 creates a \textit{schedule}. 
      In TVM, this corresponds to the order in which the loops within the 
      program are executed and, in this particular case, to 
      the three nested loops induced from the application
      of the lambda function for each $(i,j)$ entry across the reduction 
      axis~$p$.  
      By default, the schedule processes a tensor serially following a row-major order.
\item[\em Part \textsf{P4}. Specification of the loop ordering:] %
      The order in which the loops are generated with the previous schedule may not 
      match that of the baseline algorithm. 
      Lines 16-18 extract the desired axes induced from 
      the computation loops defined in Part P2, and ensure that the loops 
      follow the scheme in the basic \gemm: $i\Rightarrow j\Rightarrow p$.
\item[\em Part \textsf{P5}. Placement of the packings:] No packing occurs in the basic \gemm 
     and, therefore, this part is empty.
%
\item[\em Part \textsf{P6}. Application of fine-grain optimizations:] %
     Fine-grain optimizations such as unroll\-ing and SIMD vectorization 
     are not included in this initial version.

{\color{black}
\item[\em Part \textsf{P7}. Loop-level parallelization:] %
     Selection of the specific loop to parallelize, if necessary.    
}
\item[\em Part \textsf{P8}. Generation of code:] %
      Finally, line \textcolor{black}{36} instructs TVM to generate the code, in this case, for an LLVM backend.      
\end{description}

Other loop orderings are easily obtained using TVM by simply 
re-arranging differently the loop variables 
in line 18. (However, this is also straightforward to do in the \textcolor{black}{basic triple nested loop code written in C.})
More interestingly, this example also illustrates that code for
distinct backends can be obtained by simply changing the \texttt{target} in line 36.
Concretely, the generic {\tt llvm} target there generates code for the machine in which the generator is executed, 
but the commented lines in Part {\textcolor{black}{\textsf{P8}}
offer several examples which generate code for other architectures.
To close the discussion of this first TVM generator, note that 
it is independent of the matrix operands' data types.


\begin{figure}[tbh!]
\centering
\begin{minipage}[t]{0.9\columnwidth}
\begin{lstlisting}[language=python,alsoletter={.},deletekeywords={.sum}]
def basic_GEMM(m, n, k, dtype):
  # P1) Declare the input operands
  A = te.placeholder((m, k), name="A", dtype=dtype)
  B = te.placeholder((k, n), name="B", dtype=dtype)

  # P2) Define the operation
  p = te.reduce_axis((0, k), "p")
  C = te.compute((m, n), lambda i, j: 
                 te.sum(A[i, p] * B[p, j], 
                 axis=p), name="C")

  # P3) Prepare the schedule
  sched  = te.create_schedule(C.op)

  # P4) Specify the loop ordering (schedule i->j->p)
  (i, j) = C.op.axis
  (p,)   = C.op.reduce_axis
  sched[C].reorder(i, j, p)

  # P5) Emplace the packings
  #     No packings in this version

  # P6) Apply fine-grain optimizations
  #     No optimizations in this version

  # P7) Loop-level parallelization
  #     No parallelization in this version

  # P8) Generate code with LLVM backend
  #     A few examples for specific targets:
  #      - ARMv8a (8.2) with NEON support
  #         llvm -device=arm_cpu -mattr=+v8.2a,+fp-armv8,+neon"          
  #      - ARMv8a (8.2) with NEON and FP16 support
  #         llvm -device=arm_cpu -mattr=+v8.2a,+fp-armv8,+neon,+fp16fml  
  #      - AMD Zen2 with AVX2 support
  #         llvm -mcpu=znver2
  #      - Intel Icelake with AVX512 support
  #         llvm -mcpu=icelake-server                                    
  return tvm.build(sched, [A, B, C], target="llvm")
\end{lstlisting}
\end{minipage}
\caption{TVM generator for the basic \gemm.}
\label{lst:basic_GEMM} 
\end{figure}

\subsection{Blocking for the baseline algorithm with TVM}

Our next goal is to build a blocked algorithm that partitions the matrix operands mimicking 
the three outermost loops of the baseline algorithm (B3A2C0),
labeled as \texttt{L1}, \texttt{L2} and \texttt{L3}. 
For this purpose,  remember that 
these loops 
divide $A$ into 
blocks 
of dimension $m_c \times k_c$, 
$B$ into 
blocks 
of dimension $k_c \times n_c$, and
$C$ into 
blocks 
of dimension $m_c \times n_c$;
see Figure~\ref{fig:blis_family_Cresident} (left).
For clarity, consider the following blocked algorithm, 
which realizes the partitioning scheme in the baseline algorithm for \gemm
in order to decompose the matrix multiplication into a collection of finer-grain computations:
\vspace*{-1ex}
\begin{center}
\begin{minipage}[t]{0.8\textwidth}
\begin{lstlisting}[language=C,alsoletter={.},deletekeywords={.sum},morekeywords={},backgroundcolor=\color{white},numbers=none,frame=none]
for (jc=0; jc<n; jc+=nc) 
  for (pc=0; pc<k; pc+=kc) 
    for (ic=0; ic<m; ic+=mc) 
      for (jr=0; jr<nc; jr++) 
        for (ir=0; ir<mc; ir++) 
          for (pr=0; pr<kc; pr++) 
            C(ic+ir,jc+jr) += A(ic+ir,pc+pr) * B(pc+pc,jc+jr);
\end{lstlisting}
\end{minipage}
\end{center}


\begin{figure}[tbh!]
\centering
\begin{minipage}[t]{0.9\columnwidth}
\begin{lstlisting}[language=python,alsoletter={.},deletekeywords={.sum}]
def block_GEMM_B3A2C0(m, n, k, mc, nc, kc, dtype):
  # P1) Declare the input operands
  #     As in basic_GEMM. Omitted for brevity

  # P2) Define the operation with blocking
  #     As in basic_GEMM. Omitted for brevity

  # P3) Prepare the schedule with blocking
  sched = te.create_schedule(C.op)
  ic, jc, \
  ir, jr = sched[C].tile(C.op.axis[0], 
                         C.op.axis[1], mc, nc)
  pc, pr = sched[C].split(p, factor=kc)

  # P4) Specify the loop ordering (schedule as in baseline/B3A2C0)
  sched[C].reorder(jc, pc, ic, jr, ir, pr)

  # P5) Emplace the packings
  #     No packing available in this version

  # P6) Apply fine-grain optimizations
  #     No optimizations applied

  # P7) Loop-level parallelization
  #     No parallelization in this version

  # P8) Generate code with LLVM backend
  return tvm.build(sched, [A, B, C], target="llvm") 
\end{lstlisting}
\end{minipage}
\caption{TVM generator for \gemm mimicking the blocking scheme 
of the baseline algorithm.}
\label{lst:blocked_GEMM}
\end{figure}

Figure~\ref{lst:blocked_GEMM} 
displays a TVM generator that produces a code for \gemm that will
perform the computation adopting the same blocking scheme.
Compared with the TVM generator for the basic \gemm in Figure~\ref{lst:basic_GEMM}, 
the routine 
presents the following differences:
\begin{description}
\item[\em Part \textsf{P0}. Parameter list:] 
In addition to the parameters from the basic \gemm, the function receives the blocking parameters {\tt mc, nc, kc}.
\item[\em Part P3. Preparation of the schedule (with blocking):] In preparation for the operands' 
  tiling, lines 9--13 prompt the sought-after splittings of the problem dimensions $m$, $n$, $k$. 
\item[\em Part P4. Specification of the loop reordering:]  Line 16 specifies a loop ordering which matches that of the baseline algorithm for \gemm:
  $j_c \Rightarrow
   p_c \Rightarrow
   i_c \Rightarrow
   j_r \Rightarrow
   i_r \Rightarrow
   p_r$.
\end{description}
Here and in the following, for brevity, we will omit the parts that remain the same with
respect to the prior generators.

The code for \gemm produced by TVM using the algorithm in
Figure~\ref{lst:blocked_GEMM} 
differs from the high performance realizations of \gemm in libraries such 
as GotoBLAS, BLIS and OpenBLAS, 
in two important aspects: 
\begin{enumerate}
\item There are no packing routines that re-arrange the contents of the input matrices
      $A,B$ into buffers in order to streamline the operation of the micro-kernel.
\item The innermost loop does 
   not formulate the computation in terms of an outer product that updates a small 
   $m_r \times n_r$ micro-tile $C_r$.
   Instead, the TVM generator produces a code which decomposes this last computation into fine-grain
   multiply-add operations involving individual elements of $A, B, C$.
\end{enumerate}
\textcolor{black}{In the next subsection we deal with the first issue while the second one
is discussed in Section~\ref{sec:tvm-microkernels}.}

\subsection{Packing for the baseline algorithm with TVM}
\label{subsec:packing}

\begin{figure}[th]
\centering
\begin{minipage}[t]{0.9\columnwidth}
\begin{lstlisting}[language=python,alsoletter={.},deletekeywords={.sum}]
def packed_GEMM_B3A2C0(m, n, k, mc, nc, kc, mr, nr, dtype): 
  # P1) Declare the input operands
  #     As in block_GEMM_B3A2C0. Omitted for brevity
  
  # P2) Define the operation with (blocking and) packing

  #     Ac: 4D tensor with packed version of A
  Ac = te.compute((math.ceil(k/kc),
                   math.ceil(m/mr), kc, mr), 
                   lambda i, j, q, r: 
                   A[j * mr + r, i * kc + q], 
                   name="Ac")

  #     Bc: 4D tensor with packed version of B
  Bc = te.compute((math.ceil(k/kc), 
                   math.ceil(n/nr), kc, nr), 
                   lambda i, j, q, r: 
                   B[ i * kc + q, j * nr + r ], 
                   name="Bc")
  
  p = te.reduce_axis((0, k), "p")
  C = te.compute((m, n), lambda i, j:
         te.sum(Ac[p//kc, i//mr, 
                   tvm.tir.indexmod(p, kc), 
                   tvm.tir.indexmod(i, mr)] *
                Bc[p//kc, j//nr,
                   tvm.tir.indexmod(p, kc),
                   tvm.tir.indexmod(j, nr)],
                axis=p), name="C")
    
  # P3) Prepare the schedule with blocking
  #     As in block_GEMM_B3A2C0. Omitted for brevity

  # P4) Specify the loop ordering (schedule as in baseline/B3A2C0)
  #     As in block_GEMM_B3A2C0. Omitted for brevity

  # P5) Emplace the packings
  sched[Bc].compute_at(sched[C], pc)
  sched[Ac].compute_at(sched[C], ic) 

  # P6) Apply fine-grain optimizations
  #     No optimizations applied

  # P7) Loop-level parallelization
  #     No parallelization in this version

  # P8) Generate code with LLVM backend
  return tvm.build(sched, [A, B, C], target="llvm")
\end{lstlisting}
\end{minipage}
\caption{TVM generator for \gemm mimicking the (blocking scheme and)
         \textit{packing} of
        the baseline algorithm.}
\label{lst:packed_GEMM}
\end{figure}

Figure~\ref{lst:packed_GEMM} refines the TVM generator for the blocked algorithm 
to include
hints to TVM in order to produce a specialized version that packs $A$ and $B$ 
into two buffers, $A_c$ and $B_c$,
with the same layout as that present in the 
baseline algorithm; see Figure~\ref{fig:blis_packing}.
We note the following differences with respect to the 
TVM generator for the blocked algorithm: 
\begin{description}
\item[\em Part \textsf{P0}. Parameter list:] 
     The new routine receives two additional parameters: {\tt mr, nr}. 
\item[\em Part P2. Definition of the operation and packing schemes for $A$ and $B$:] %
      Lines 8--12 de\-cla\-re a 4D TVM tensor {\tt Ac} 
      which will hold a packed version of the complete operand $A$.
      {\tt Ac} is instantiated as a (2D) matrix of small (packed) 
      matrices ({\em micro-panels} of $A$), each of dimension
      $m_r \times k_c$; see the red micro-panel labeled as $A_r$
      in Figure~\ref{fig:blis_packing}.
      \texttt{Ac} will potentially store {\em all} packed micro-panels
      of $A$ that will be generated during the iterations of loop {\tt L3} 
      in Figure~\ref{fig:blis_family_Cresident}, and hence its global dimension 
      initially matches that of $A$. {\color{black}However, depending on the placement of the packing specified by the user, TVM will evaluate the potential reuse level of the buffer(s) and will accordingly adapt the buffer dimensions. (This will be refined in the following discussion
      of part P5.)}
      The lambda function in lines 10--11 specifies
      the desired packing scheme by mapping the elements of {\tt Ac}
      to their counterparts in  $A$.
      Concretely, the four arguments 
      (\texttt{i, j, q, r}) of the function determine how 
      the element \texttt{Ac[i,j,q,r]} is extracted from $A$
      conformally with the packing scheme for the baseline algorithm. 
      
      Lines 15--19, involving the tensor {\tt Bc} and the matrix $B$,
      play an analogous role to that described previously
      for {\tt Ac} and $A$.
      
      Lines 21--29 define the operation to be computed in terms of the tensors {\tt Ac} and {\tt Bc}. 
      \textcolor{black}{Here {\tt Ac} is transposed (with respect to {\tt A}), which is necessary to ensure
      that TVM generates a code that accesses the entries of {\tt Ac} with unit stride.}
      Also the computation is defined in terms of a reduction ({\tt sum}) across
      the dimension {\tt k}.
       
\item[\em Part P5. Placement of the packings:] %
      Lines 38--39 place the computation of tensors {\tt Bc} and {\tt Ac} at
      the desired points of the loop nesting:
      Concretely, inside the loop indexed by \texttt{pc} for {\tt Bc} and the loop indexed by \texttt{ic} for {\tt Ac}; 
      see the algorithm on the left of Figure~\ref{fig:blis_family_Cresident}.
      {\color{black} As a result of combining the correct placement of the tensor buffers, loop ordering,
      and access pattern in the blocked computation of {\tt C}, the packed tensors
      {\tt Ac} and {\tt Bc} do not need to have the same size as {\tt A} and {\tt B}, respectively. Instead, they are computed at each iteration 
      of loop {\tt L3} (indexed by {\tt ic}) and loop {\tt L2} (indexed by {\tt pc}).} 
      In the latter case, for example, the reuse pattern
      of elements for {\tt Ac} guides the compiler to reduce the dimension of the packed
      buffer from $m \times k$ to $m_c \times k_c$ only (which matches the dimension of the 
      buffer $A_c$ in Figure \ref{fig:blis_packing}). 
      An analogous comment applies to matrix $B$ and its packed tensor counterpart.
\end{description}

The purpose of the packing operations 
is to favor a higher number of cache hits when accessing the data in the L2 and L1 caches
from the micro-kernel.
The packing introduces a certain overhead,
but this is usually low
because the packed data is  re-utilized multiple times~\cite{BLIS4}. 
As a side note, let us comment that, depending on the problem dimensions
($m,n,k$) and the loop strides 
($m_c,n_c,k_c$), in some cases the costs of the packing operations may exceed their benefits.
Fortunately, with TVM we can easily eliminate any of the packing operations.
For example, in the 
baseline algorithm in 
\textcolor{black}{Figure~\ref{lst:packed_GEMM}},
the two packing operations can be eliminated by 
\textcolor{black}{
1) removing lines~8-19; changing the references
from {\tt Ac} and {\tt Bc} to {\tt A} and {\tt B} in lines~23, 26;
and 2) adapting accordingly the indexing on {\tt A} and {\tt B} to accommodate the dimensionality
modification from {\tt Ac} and {\tt Bc} (from 4D to 2D).}

The intermediate representation (IR) produced by TVM 
using the generator for the packed realization of \gemm 
can be described as follows:
\begin{center}
\begin{minipage}[t]{0.6\textwidth}
\begin{lstlisting}[language=C,alsoletter={.},deletekeywords={.sum},morekeywords={},backgroundcolor=\color{white},numbers=none,frame=none]
for (jc=0; jc<n; jc+=nc) {         // Loop L1 
  for (pc=0; pc<k; pc+=kc) {       // L2
    for (i=0; ...) {               // Packing Bc
      for (j=0; ...)
        for (q=0; ...)
          for (r=0; ...)
            Bc[...] = B[...];
    }       
    for (ic=0; ic<m; ic+=mc) {     // L3
      for (i=0; ...) {             // Packing Ac
        for (j=0; ...)
          for (q=0; ...)
            for (r=0; ...)
              Ac[...] = A[...];
      }     
      for (jr=0; jr<nc; jr++)     // L4
        for (ir=0; ir<mc; ir++)   // L5
          for (pr=0; pr<kc; pr++) // L6
            C[...] += Ac[...] * Bc[...];
} } } 
\end{lstlisting}
\end{minipage}
\end{center}
(Here we omitted a few indexing details and introduced some formatting to ease the interpretation.)

This takes us one step forward toward automatically generating 
a high performance realization of \gemm that mimics 
the baseline algorithm for \gemm by removing one of the caveats
that appeared in the previous blocking algorithm: the lack of packing routines. In the next section
we address the second missing detail: 
\textit{the formulation of the innermost loop of the algorithm as a micro-kernel that performs 
an outer product.}

\section{Automatic Generation of High Performance Micro-kernels for \gemm}
\label{sec:tvm-microkernels}

The final stage in our journey to obtain a high performance realization of \gemm has the 
goal of integrating the automatic generation of high performance micro-kernels 
within the TVM routine for \gemm. 

\subsection{Micro-kernel with \Cresident}

Starting from the TVM generator 
in Figure~\ref{lst:packed_GEMM}, we introduce two main changes
with the result shown in Figure~\ref{lst:packed_GEMM_w_ukernel}: 
\begin{description}
\item[\em Part P3. Preparation of the schedule (with additional blocking):]  The $m_c,n_c$ dimensions 
of the problem are respectively partitioned using factors $m_r, n_r$, 
in order to expose the loops that will operate with the individual elements of the 
micro-panel $C_r$; see lines 13--14.

\item[\em Part P4. Specification of the loop ordering:]  Lines 17--18 place
the newly created loops for the micro-kernel, {\tt it} and {\tt jt},
within the global schedule.

\end{description}


\begin{figure}
\centering
\begin{minipage}[t]{0.9\columnwidth}
\begin{lstlisting}[language=python,alsoletter={.},deletekeywords={.sum}]
def packed_GEMM_B3A2C0_ukernel(m, n, k, mc, nc, kc, mr, nr, dtype):
  # P1), P2) Declare the input operands and define the operation
  #          As in packed_GEMM_B3A2C0. Omitted for brevity

  # P3) Prepare the schedule with blocking
  sched = te.create_schedule(C.op)
  ic, jc, \
  ir, jr = sched[C].tile(C.op.axis[0], 
                         C.op.axis[1], mc, nc)
  pc, pr = sched[C].split(p, factor=kc)

  #     Expose loops inside micro-kernel
  ir, it = sched[C].split(ir, factor=mr)
  jr, jt = sched[C].split(jr, factor=nr)

  # P4) Specify the loop ordering (schedule as in baseline/B3A2C0)
  sched[C].reorder(jc, pc, ic, 
                   jr, ir, pr, it, jt)

  # P5) Emplace the packings
  #     As in packed_GEMM_B3A2C0

  # P6) Apply fine-grain optimizations
  #     No optimizations applied
  
  # P7) Loop-level parallelization
  #     No parallelization in this version

  # P8) Generate code with LLVM backend
  return tvm.build(sched, [A, B, C], target="llvm")
\end{lstlisting}
\end{minipage}
 \caption{TVM generator for \gemm mimicking (the blocking and packing of)
        the baseline algorithm, and \textit{integrating the optimized micro-kernel with \Cresident}.}
\label{lst:packed_GEMM_w_ukernel}
\end{figure}

\subsection{Fine-grain optimizations for high performance \textcolor{black}{and parallelization}\label{sec:microkernels_opt}}

The TVM generator can explore relevant optimizations at the micro-kernel level 
which pursue goals that are similar to those of the low-level optimizations 
applied by the expert Assembly programmer.
By off-loading these optimizations to TVM, 
in exchange for less flexibility, there is a considerably more 
reduced programming effort and higher performance portability across different processor
architectures.

Starting from the TVM generator in Figure~\ref{lst:packed_GEMM_w_ukernel}, the enhanced alternative in Figure~\ref{lst:opt_GEMM_w_ukernel} 
illustrates the necessary additions to introduce 
some conventional optimization techniques:
\begin{description}
\item[\em Part P6. Fine-grain optimizations:] This invokes TVM methods to address three types of optimizations: 
\begin{itemize}
    \item {\em SIMD vectorization:} The generator includes vectorization of three different stages of the blocked algorithm to leverage the SIMD units
    of the architectures: 
    micro-kernel computation (lines 8--10); packing $A$ into $A_c$ (lines 13--16); 
    and packing $B$ into $B_c$ (lines 19--22). 
    The basic scheme remains the same for all of them:
    the innermost loop ($jt$ for the former and $l$ in latter two) 
    is split to expose an additional inner loop, using a split factor ({\tt lanesize}) that matches the
    SIMD width of the target architecture. For example, for 32-bit floating point data,
    {\tt lanesize} is set to 4 for ARMv8a NEON (128-bit SIMD registers) and 16 for 
    Intel AVX512 (512-bit SIMD registers). In general, in our experiments we will employ the exact number of elements for a SIMD register, or an integer multiple of it. 
    From the newly exposed loops, the outer one is unrolled 
    and the inner is vectorized 
    ({\tt unroll} and {\tt vectorize} methods, respectively) 
    This produces a code comprising vector instructions for the desired target
    processor.
    \item {\em Prefetching and vector loads within the micro-kernel:}  {\tt ac} and {\tt bc} are tensors defined as read-only copies of {\tt Ac} and {\tt Bc}. 
    The purpose of creating these artifacts is two-fold: first, it helps the compiler to
    introduce software-prefetching instructions in order to pre-load the micro-panels 
    of {\tt Ac} and {\tt Bc} that will be used throughout the computation of the micro-kernel; 
    second, it instructs the compiler to use vector registers in order 
    to store the active parts of {\tt ac} and {\tt bc} in the micro-kernel, 
    and to use vector instructions to load them from {\tt Ac} and {\tt Bc}.
Diving into the details, the construction, re-use pattern and computation point (loop indexed by \texttt{pr}) of \texttt{ac} and \texttt{bc} induce buffers that store the strictly necessary micro-panels of \texttt{Ac} and \texttt{Bc} used within the computation of each iteration of loop \texttt{pr} within the micro-kernel, that is, $m_r \times 1$ for 
\texttt{ac}, and $n_r \times 1$ for \texttt{bc}. Splitting the innermost loop (\texttt{l}) using a factor that matches {\tt lanesize}, unrolling and vectorizing it yield vector instructions to load the micro-panels of \texttt{Ac} and \texttt{Bc} into $m_r / lanesize$ and $n_r / lanesize$ vectors. 
\end{itemize}
{\color{black}
\item[\em Part P7. Loop-level parallelization:] TVM also allows exploiting loop parallelism via multi-threading.  Line 41 shows how to instruct TVM to split the iteration space for a given loop across the active threads. The selection of the optimal loop 
to parallelize depends on the problem dimensions and target architecture specifications; 
see~\cite{BLIS3} for a detailed discussion. 
}

\end{description}


\textcolor{black}{Figure~\ref{lst:microkernel_4x4} displays two examples of the 
assembly code generated by TVM: An 
$m_r \times n_r = 4 \times 4$ micro-kernel for an ARMv8a ISA 
(instruction set architecture)
with NEON vector instructions and unrolling factor of 4 on the left;
and an $m_r \times n_r = 4 \times 16$ micro-kernel for an x86 ISA with AVX512 
vector instructions (on the right).
In both cases, for brevity, we only include the instructions that 
\textcolor{black}{are comprised by}
the reduction loop of the micro-kernel.
The codes were obtained using the generator for the optimized B3A2C0 
variant depicted in Figure~\ref{lst:opt_GEMM_w_ukernel}, selecting 
the target as 
{\tt llvm -device=arm\_cpu -mattr=+v8.2a,+fp-armv8,+neon} for the ARMv8a architecture,
and
{\tt llvm -mcpu=icelake-server} for the x86 architecture. 
The codes share a similar structure, 
but present subtle differences in terms of 
syntax and vectorization strategy: The ARMV8a codes rely on vector fused multiply-add instructions
with a single element of $B$ as a source (e.g. {\tt fmla v3.4s,v5.4s,v4.s[0]});
in contrast, 
the x86 counterpart operates on a broadcast operation and vector fused multiply-add strategy 
({\tt vbroadcastss \%xmm4,\%zmm8} followed by \texttt{vfmadd213ps \%zmm3,\%zmm6,\%zmm8} ).
}

\begin{figure}
\centering
\begin{minipage}[t]{0.9\columnwidth}
\begin{lstlisting}[language=python,alsoletter={.},deletekeywords={.sum}]
def opt_GEMM_B3A2C0_ukernel(m, n, k, mc, nc, kc, mr, nr, lanesize, dtype):
  # P1), P2) P3) P4) P5) as in packed_GEMM_B3A2C0_ukernel
  #     Omitted for brevity

  # P6) Apply fine-grain optimizations

  #     6.1. Vectorization of the microkernel
  jto,jt = sched[C].split(jt, factor=lanesize)
  sched[C].unroll(jto)
  sched[C].vectorize(jt)

  #     6.2. Vectorization of the packing of A
  x,y,z,l = Ac.op.axis
  l,ll = sched[Ac].split(l, factor=lanesize)
  sched[Ac].unroll(l)
  sched[Ac].vectorize(ll)

  #     6.3. Vectorization of the packing of B
  x,y,z,l = Bc.op.axis
  l,ll = sched[Bc].split(l, factor=lanesize)
  sched[Bc].unroll(l)
  sched[Bc].vectorize(ll)
  
  #     6.4. Prefetching
  ac = sched.cache_read(Ac, "global", readers=[C])
  bc = sched.cache_read(Bc, "global", readers=[C])
  
  sched[ac].compute_at(sched[C],pr)
  x,y,z,l = ac.op.axis
  l,ll = sched[ac].split(l, factor=lanesize)
  sched[ac].unroll(l)
  sched[ac].vectorize(ll)
  
  sched[bc].compute_at(sched[C],pr)
  x,y,z,l = bc.op.axis
  l,ll = sched[bc].split(l, factor=lanesize)
  sched[bc].unroll(l)
  sched[bc].vectorize(ll)
  
  # P7) Loop-level parallelization on ic
  sched[C].parallel(ic)

  # P8) Generate code with LLVM backend
  return tvm.build(sched, [A, B, C], 
                   target="llvm")
\end{lstlisting}
\end{minipage}
 \caption{TVM generator for \gemm mimicking (the blocking and packing of)
        the baseline algorithm, integrating both the optimized micro-kernel with \Cresident and
        {\em fine-grain optimizations and a loop-level parallelization of loop {\em ic}.}}
\label{lst:opt_GEMM_w_ukernel}
\end{figure}

\begin{figure}
\centering
\begin{minipage}[t]{0.40\columnwidth}
\begin{lstlisting}[language={[x86masm]Assembler},alsoletter={.},keywords={fmla,add,ldr,b.ne,adds}]
.LBB1_5:
  add  x11, x9, x10
  add  x12, x8, x10
  ldr  q4, [x11, #15952]
  ldr  q5, [x12, #15968]
  ldr  q6, [x11, #15968]
  ldr  q7, [x12, #15984]
  ldr  q16, [x11, #15984]
  ldr  q17, [x12, #16000]
  fmla v3.4s, v5.4s, v4.s[0]
  fmla v2.4s, v5.4s, v4.s[1]
  fmla v1.4s, v5.4s, v4.s[2]
  fmla v0.4s, v5.4s, v4.s[3]
  ldr  q4, [x11, #16000]
  ldr  q5, [x12, #16016]
  fmla v3.4s, v7.4s, v6.s[0]
  fmla v2.4s, v7.4s, v6.s[1]
  fmla v1.4s, v7.4s, v6.s[2]
  fmla v0.4s, v7.4s, v6.s[3]
  fmla v3.4s, v17.4s, v16.s[0]
  fmla v2.4s, v17.4s, v16.s[1]
  fmla v1.4s, v17.4s, v16.s[2]
  fmla v0.4s, v17.4s, v16.s[3]
  adds x10, x10, #64
  fmla v3.4s, v5.4s, v4.s[0]
  fmla v2.4s, v5.4s, v4.s[1]
  fmla v1.4s, v5.4s, v4.s[2]
  fmla v0.4s, v5.4s, v4.s[3]
  b.ne .LBB1_5
\end{lstlisting}
\end{minipage}
\hfill
\begin{minipage}[t]{0.50\columnwidth}
\begin{lstlisting}[language={[x86masm]Assembler},alsoletter={.},keywords={vmovaps,vmovups,vbroadcastss,jne,cmpq,addq,vfmadd213ps,vfmadd231ps,vmovshdup,vpermilps,vbroadcastsd}]
.LBB5_12:
  vmovaps       -16(%rdx,%r13), %xmm4
  vmovaps       (%rdx,%r13), %xmm5
  vmovups       -64(%rsi,%r13,4), %zmm6
  vmovups       (%rsi,%r13,4), %zmm7
  vbroadcastss  %xmm4, %zmm8
  vfmadd213ps   %zmm3, %zmm6, %zmm8
  vmovshdup     %xmm4, %xmm3
  vbroadcastsd  %xmm3, %zmm9
  vfmadd213ps   %zmm2, %zmm6, %zmm9
  vpermilps     $234, %xmm4, %xmm2
  vbroadcastsd  %xmm2, %zmm10
  vfmadd213ps   %zmm1, %zmm6, %zmm10
  vpermilps     $239, %xmm4, %xmm1
  vbroadcastsd  %xmm1, %zmm4
  vfmadd213ps   %zmm0, %zmm6, %zmm4
  vbroadcastss  %xmm5, %zmm3
  vfmadd213ps   %zmm8, %zmm7, %zmm3
  vmovshdup     %xmm5, %xmm0
  vbroadcastsd  %xmm0, %zmm2
  vfmadd213ps   %zmm9, %zmm7, %zmm2
  vpermilps     $234, %xmm5, %xmm0
  vbroadcastsd  %xmm0, %zmm1
  vfmadd213ps   %zmm10, %zmm7, %zmm1
  vpermilps     $239, %xmm5, %xmm0
  vbroadcastsd  %xmm0, %zmm0
  vfmadd213ps   %zmm4, %zmm7, %zmm0
  addq          $32, %r13
  cmpq          $2048, %r13
  jne           .LBB5_12
\end{lstlisting}
\end{minipage}
 \caption{Assembly codes for the reduction loop (traversing $k_c$) of micro-kernels with \Cresident 
 automatically generated by TVM for single precision. Left: ARMv8a assembly code with NEON vector instructions for a $4 \times 4$ micro-kernel with a loop unrolling technique with a factor of 4. Right: x86 assembly code with AVX512 vector instructions for a $4 \times 16$ micro-kernel.}
\label{lst:microkernel_4x4}
\end{figure}


\section{Other Members in the Family of Blocked Algorithms for \gemm}
\label{sec:otherfamily}


One of the advantages of TVM lies in that producing code for other blocked algorithms of the \gemm family
only requires small changes in the generator routines
in order to accommodate the proper loop ordering, blocking and packing schemes. This is described 
in this section.


\subsection{Blocking and packing for variants with $C$ in the L2 cache}


\begin{figure}[th]
\centering
\begin{minipage}[t]{0.9\columnwidth}
\begin{lstlisting}[language=python,alsoletter={.},deletekeywords={.sum}]
def packed_GEMM_A3C2B0(m, n, k, mc, nc, kc, kr, nr, dtype):
  # P1) Declare the input operands
  #     As in packed_GEMM_B3_A2C0. Omitted for brevity

  # P2) Define the operation with blocking and packing

  #     Ac: 4D tensor with packed version of A
  Ac = te.compute((math.ceil(m/mc), 
                   math.ceil(k/kr), mc, kr), 
                   lambda i, j, q, r:
                   A[i * mc + q, j * kr + r],
                   name="Ac")

  #     Bc: 4D tensor with packed version of B
  Bc = te.compute((math.ceil(k/kr), 
                   math.ceil(n/nr), kr, nr), 
                   lambda i, j, q, r: 
                   B[i * kr + q, j * nr + r],
                   name="Bc")
    
  p = te.reduce_axis((0, k), "p")
  C = te.compute((m, n), lambda m, n: 
         te.sum(Ac[m // mc, p // kr, 
                   tvm.tir.indexmod(m, mc),
                   tvm.tir.indexmod(p, kr)] * 
                Bc[p // kr, n // nr, 
                   tvm.tir.indexmod(p, kr), 
                   tvm.tir.indexmod(n, nr)], 
                axis=p), name="C")

  # P3) Prepare the schedule with blocking
  sched = te.create_schedule(C.op)

  cic, cjc, \
  cir, cpr = sched[C].tile(C.op.axis[0], 
                           C.op.axis[1], 
                           mc, nc)

  Cc = sched.cache_write(C, "global")

  ic, jc, \
  ir, jr = sched[Cc].tile(Cc.op.axis[0], 
                          Cc.op.axis[1], 
                          mc, nc)
  pc, pr = sched[Cc].split(p, factor=kc)

  # P4) Specify loop ordering (schedule as in A3C2B0)
  sched[Cc].reorder(ic, pc, jc, pr, jr, ir)
    
  # P5) Define packing placement
  sched[Cc].compute_at(sched[C], cic)
  sched[Ac].compute_at(sched[Cc], pc)
  sched[Bc].compute_at(sched[Cc], ir)
    
  # P6) Apply fine-grain optimizations
  #     No fine-grain optimizations applied.

  # P7) Loop-level parallelization
  #     No parallelization in this version

  # P8) Code generation
  return tvm.build(sched, [A, B, C], target="llvm")
\end{lstlisting}
\end{minipage}
\caption{TVM generator for \gemm mimicking the blocking and packing schemes of
        the A3C2B0 algorithm.}
\label{lst:packed_A3C2B0}
\end{figure}

Transforming the TVM generator for the baseline algorithm, 
with a micro-tile of \Cresident 
, (left column 
in Figure~\ref{fig:blis_family_Cresident})
into that with a block of $C$ in the L2 cache and a micro-tile of 
\Bresident 
(A3C2B0, right column in Figure~\ref{fig:blis_family_CL2}) 
requires a number of changes in the TVM generator, shown in 
Figure~\ref{lst:packed_GEMM}, resulting in the variant displayed 
in Figure~\ref{lst:packed_A3C2B0}:

\begin{description}
\item[\em Part \textsf{P0}. Parameter list:]
This generator includes {\tt kr} as a parameter
      (instead of {\tt mr}).
\item[\em Part P2. Definition of the operation and packing schemes for $A$ and $B$:] %
   Lines 15--19 modify the dimensions of the {\tt Bc} tensor (this version moves $B$ to registers) so that the size of each submatrix (micro-panel) in {\tt Bc} is $k_r \times n_r$. 
   
\item[\em Part P3. Preparation of the schedule:] %
   Lines 34--37 apply tiling over $C$ in pre\-pa\-ra\-tion for the placement of the packing into the buffer {\tt Cc}.
   In addition, line~39 creates a buffer for reading and writing the matrix $C$. From that point, the scheduler operates with the object {\tt Cc}. \textcolor{black}{This change is crucial for the algorithms where $C$ resides in either the L2 or L3 caches. Concretely, these variants leverage
   {\tt cache\_write} to induce a copy from matrix $C$ to the  object {\tt Cc} 
   as well as to restore the result of the micro-kernel, temporarily in {\tt Cc}, back into $C$.}

\item[\em Part P4. Specification of the loop ordering:] %
Line 48 specifies a loop order that matches that of the algorithm A3C2B0. 

\item[\em Part P5. Placement of the packings:] %
 Lines 51--53 set the buffer for {\tt Cc} and the {\tt Ac} and {\tt Bc} tensors into the appropriate loops. 
    

\end{description}

The TVM generator in Figure~\ref{lst:packed_A3C2B0} does not include fine-grain optimizations,
yet identical techniques as those introduced in Section~\ref{sec:microkernels_opt} 
apply for this member of the family.
Also, the TVM generator to obtain variant \textcolor{black}{B3C2A0} can be easily derived following the same principles.

\subsection{Blocking and packing for variants with $C$ in the L3 cache}


{\color{black} The remaining two blocked algorithmic variants for \gemm
in the family store a block of matrix $C$ 
in the L3 cache throughout the computation.
Starting from the TVM generator for variant A3C2B0 in Figure~\ref{lst:packed_A3C2B0}, 
the following changes are introduced to obtain the missing TVM generator, for variant C3A2B0, displayed in Figure~\ref{lst:packed_C3A2B0}:
\begin{description}
 \item[\em Part P4. Specification of the loop ordering:] %
 Line 12 enforces a loop ordering which mimics
 that of the algorithm in Figure~\ref{fig:blis_family_CL2} (right).
    \item[\em Part P5. Placement of the packings:] %
 The buffer $C_c$ does not need to be bound to any loop because it belongs to the outer structure.
 The packings of {\tt Ac} and {\tt Bc} are placed in the same loops as in the version with $C$ in the L2 cache (see lines 15--16).
\end{description}
%
Obtaining the last variant, C3B2A0, is direct and, therefore, 
its discussion is omitted for brevity.

}

\begin{figure}[th]
\centering
\begin{minipage}[t]{0.9\columnwidth}
\begin{lstlisting}[language=python,alsoletter={.},deletekeywords={.sum}]
def packed_GEMM_C3A2B0(m, n, k, mc, nc, kc, kr, nr, dtype):
  # P1) Declare the input operands
  #     As in packed_GEMM_A3C2B0. Omitted for brevity

  # P2) Define the operation with blocking and packing
  #     As in packed_GEMM_A3C2B0. Omitted for brevity

  # P3) Prepare the schedule with blocking
  #     As in packed_GEMM_A3C2B0. Omitted for brevity

  # P4) Specify loop ordering (schedule as in C3A2B0)
  sched[Cc].reorder(jc, ic, pc, jr, pr, ir)

  # P5) Emplace the packings
  sched[Ac].compute_at(sched[Cc], pc)
  sched[Bc].compute_at(sched[Cc], ir)
 
  # P6) Apply fine-grain optimizations
  #     No fine-grain optimizations applied.

  # P7) Loop-level parallelization
  #     No parallelization in this version

  # P8) Code generation
  return tvm.build(sched, [A, B, C], target="llvm")
\end{lstlisting}
\end{minipage}
\caption{TVM generator for \gemm mimicking the blocking and packing schemes of
        the C3A2B0 algorithm.} 
\label{lst:packed_C3A2B0}
\end{figure}

\section{Experimental Results}
\label{sec:experiments}

In this section, we provide strong experimental evidence that 
\textrm{the TVM-based approach to automatically 
generate routines for the matrix multiplication 
paves an almost effortless road toward experimenting with a rich variety of 
algorithms, micro-kernels and parallelization/optimization 
options that offer fair performance in a number of scenarios}.  

\subsection{General setup}

Unless otherwise explicitly stated, 
the experiments in this section were carried out 
using a single core of
the NVIDIA Carmel processor (ARM v8.2) embedded on
an NVIDIA Jetson AGX Xavier board,
using IEEE 32-bit floating point arithmetic (FP32). 
In order to reduce variability, the processor frequency was fixed to
\textcolor{black}{2.3}~GHz, the threads were bound to the hardware cores,
and the experiments were repeated a large number of times reporting 
in the following the average results for each experiment.
In general, performance is measured in terms of billions of 
arithmetic operations per second, abbreviated as GFLOPS when 
operating with floating point arithmetic and GIOPS for integer arithmetic.

Also, unless explicitly stated, we target the baseline algorithm
for \gemm with a micro-kernel that operates with
\Cresident. 
For reference, when possible
we include in the evaluation the results obtained from up-to-date realizations of the
\gemm kernel in libraries such as BLIS
(version \textcolor{black}{v0.8.1}),
OpenBLAS (version \textcolor{black}{v0.3.19}),
and 
the ARM Performance Libraries (ARMPL, version \textcolor{black}{v21.1}).

The dataset for the experimentation includes two types of \gemm problems:
large square matrices versus highly ``rectangular'' problems. 
Given the current interest in DL inference, 
the dimensions of the 
latter are selected as those that result from applying the 
\imcol transform~\cite{Che06} 
to cast the convolution layers in the ResNet50 v1.5 deep neural 
network (DNN) model in terms of a \gemm. The ``batch'' size for the inference scenario is set 
to 128 samples.
\textcolor{black}{As some layers share the same parameters, resulting in \gemm problems 
of the same dimensions, 
we report the results for those only once (per layer type); see Table~\ref{table:resnet50}. In the following, we will use ``layer'' to refer to ``layer type'', since we are only interested in the ``layer dimensions''.}  

\begin{table}[]
\vspace*{2ex}
\centering
\footnotesize
\caption{Dimensions of the \gemm resulting from applying the \imcol transform to the layers of the ResNet50 v1.5 DNN model with a batch size of 128 samples.}
\begin{tabular}{rlrrrcrlrrr}
\toprule

Layer & Layer numbers & $m$ & $n$ & $k$ & & Layer & Layer numbers & $m$ & $n$ & $k$\\ 
 type id.  & in ResNet50 v1.5 &     &     &     & &  type id. & in ResNet50 v1.5    &     &     &    \\ \cline{1-5} \cline{7-11}
~1&001 &                     1,605,632 &  ~~64 & ~147 & &
11&080 &                     ~100,352 &  ~256  & ~512 \\
~2&006 &                     ~401,408 &  ~~64  & ~~64 & &
12&083/095/105/115/125/135 & ~~25,088 &  ~256  & 2,304 \\
~3&009/021/031 &             ~401,408 &  ~~64  & ~576 & &
13&086/098/108/118/128/138 & ~~25,088 &  1,024  & ~256 \\
~4&012/014/024/034 &         ~401,408 &  ~256  & ~~64 & &
14&088 &                     ~~25,088 &  1,024  & ~512 \\
~5&018/028 &                 ~401,408 &  ~~64  & ~256 & &
15&092/102/112/122/132 &     ~~25,088 &  ~256  & 1,024 \\
~6&038 &                     ~401,408 &  ~128  & ~256 & &
16&142 &                     ~~25,088 &  ~512  & 1,024\\
~7&041/053/063/073 &         ~100,352 &  ~128  & 1,152 & &
17&145/157/167 &             ~~~6,272 &  ~512  & 4,608 \\
~8&044/056/066/076 &         ~100,352 &  ~512  & ~128 & &
18&148/160/170 &             ~~~6,272 &  2,048  & ~512 \\
~9&046 &                     ~100,352 &  ~512  & ~256 & &
19&150 &                     ~~~6,272 &  2,048  & 1,024 \\
10&050/060/070 &             ~100,352 &  ~128  & ~512 & &
20&154/164 &                 ~~~6,272 &  ~512  & 2,048 \\ \bottomrule
\end{tabular}
\label{table:resnet50}
\end{table}

\subsection{Cache configuration parameters}

The performance of the GotoBLAS2-like realizations of \gemm as well as our routines obtained with
TVM is strongly  dictated by the optimization level of the micro-kernel and
a proper selection of the cache configuration parameters $m_c,n_c,k_c$. 
The optimal values for the latter three parameters
depend on hardware features such as number of cache levels, size, set associativity, etc.,
and the specific dimensions of the micro-kernel (given by $m_r \times n_r$ for 
a \Cresident case).
Determining these values via brute force
experimentation involves an expensive search across a large 3D space,
possibly 
for each micro-kernel dimension. 


Alternatively, one can use the analytical model in~\cite{BLIS4} to select the optimal values for the
cache configuration parameters.
The advantage of this analytical approach in our particular case will be exposed in the 
next subsection, when we automatically generate, explore and evaluate 
a variety of micro-kernels, of different dimensions, 
using the TVM routine.

\subsection{Comparison of the TVM generators}

We open the experimental evaluation by illustrating the
performance boost that is obtained by incrementally 
integrating the blocking, packing, and fine-grained
optimizations described with the successive TVM generators.
For simplicity, in this initial analysis we only consider a 
square \gemm problem with dimensions given by $m=n=k=$ 2,000. 

\begin{figure}[t!]
\begin{center}
         \includegraphics[width=0.65\textwidth]{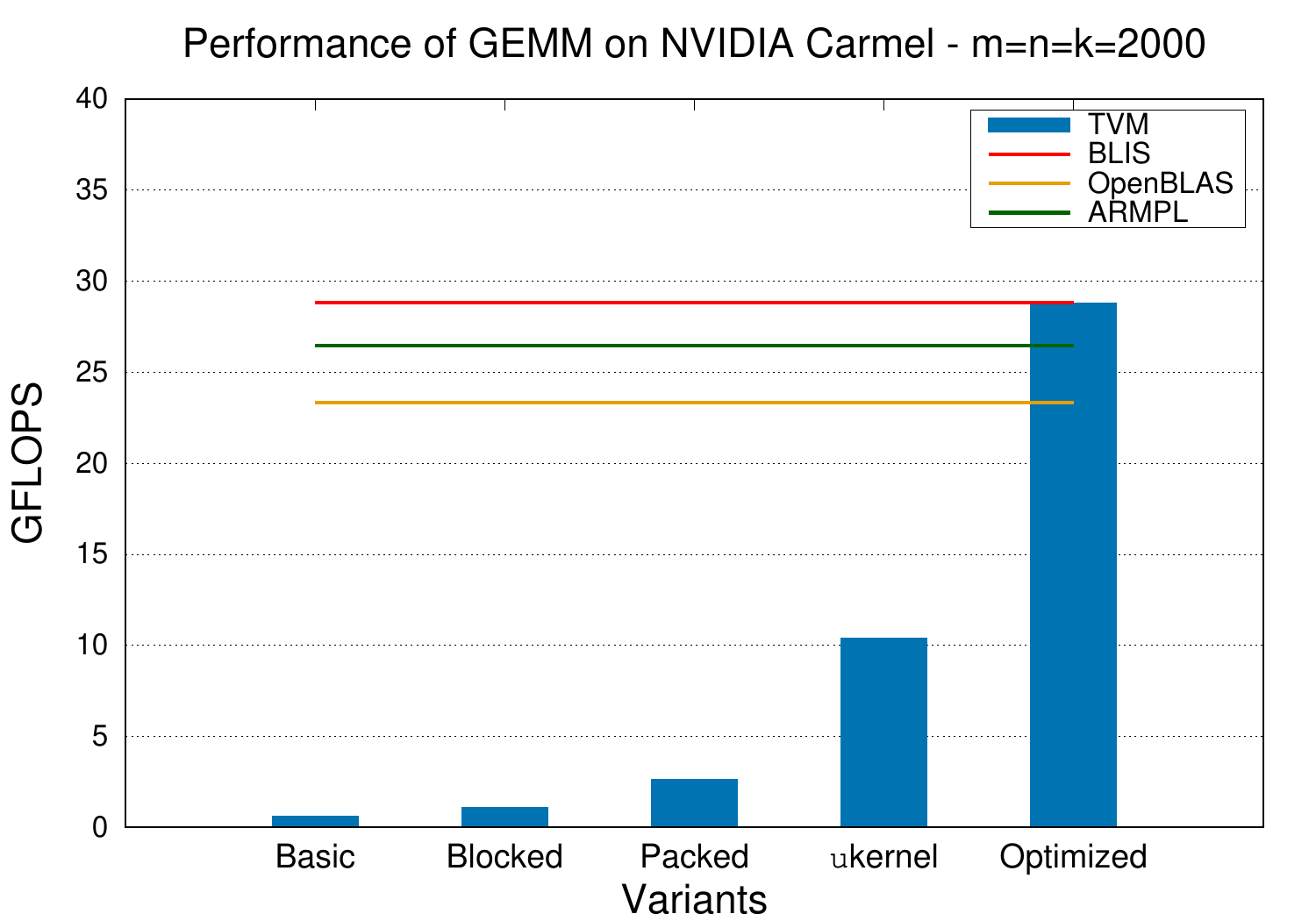}
\end{center}
\caption{Performance comparison of distinct TVM generators on a single 
         NVIDIA Carmel core for square matrices.}
\label{fig:2000_versiones} 
\end{figure}

Figure~\ref{fig:2000_versiones}
shows the results for this first experiment. The five blue bars there report the
performance attained by the \gemm routines automatically obtained with
the five TVM generators described in Sections~\ref{sec:tvm-family}
and~\ref{sec:tvm-microkernels}.
Concretely, the labels 
\textsf{\small Basic}, 
\textsf{\small Blocked}, 
\textsf{\small Packed}, 
\textsf{\small $\mu$kernel}, and 
\textsf{\small Optimized} 
in the $x$-axis respectively refer to the TVM generators
\texttt{\small basic\_GEMM}, 
\texttt{\small block\_GEMM\_B3A2C0}, 
\texttt{\small packed\_GEMM\_B3A2C0}, 
\texttt{\small packed\_GEMM\_B3A2C0\_uker\-nel}, and
\texttt{\small opt\_GEMM\_B3A2C0\_ukernel}
in Figures~\ref{lst:basic_GEMM},
\ref{lst:blocked_GEMM},
\ref{lst:packed_GEMM},
\ref{lst:packed_GEMM_w_ukernel}, and
\ref{lst:opt_GEMM_w_ukernel}.
{\color{black}
For the latter, parallelization is disabled (line 41 of Figure~\ref{lst:opt_GEMM_w_ukernel}). Hence, the reported
performance results correspond to a sequential \gemm execution.
}
For reference, the figure also displays the performance attained with the
realizations of the \gemm kernel in BLIS (red line),
OpenBLAS (yellow line), and ARMPL (green line) for this particular problem size
(on the NVIDIA Carmel core).

For this first experiment, we observe the notable performance
raise attained with the integration of the micro-kernel and the
fine-grain optimizations (rightmost two bars) as well as the fact that, for this
problem dimension (and target processor core), the
best TVM generator delivers a code for \gemm whose performance
exactly matches that of
the BLIS kernel for this operation.
This is not totally surprising since, for this initial experiment, 
we forced TVM to generate a routine with exactly the
same cache configuration parameters and micro-kernel dimensions as those used
by BLIS.
Hereafter, all our TVM results correspond to the 
version of \gemm obtained with the optimized TVM generator
\texttt{\small opt\_GEMM\_B3A2C0\_ukernel}.


\begin{figure}[thb]
\begin{center}
\includegraphics[width=0.48\textwidth]{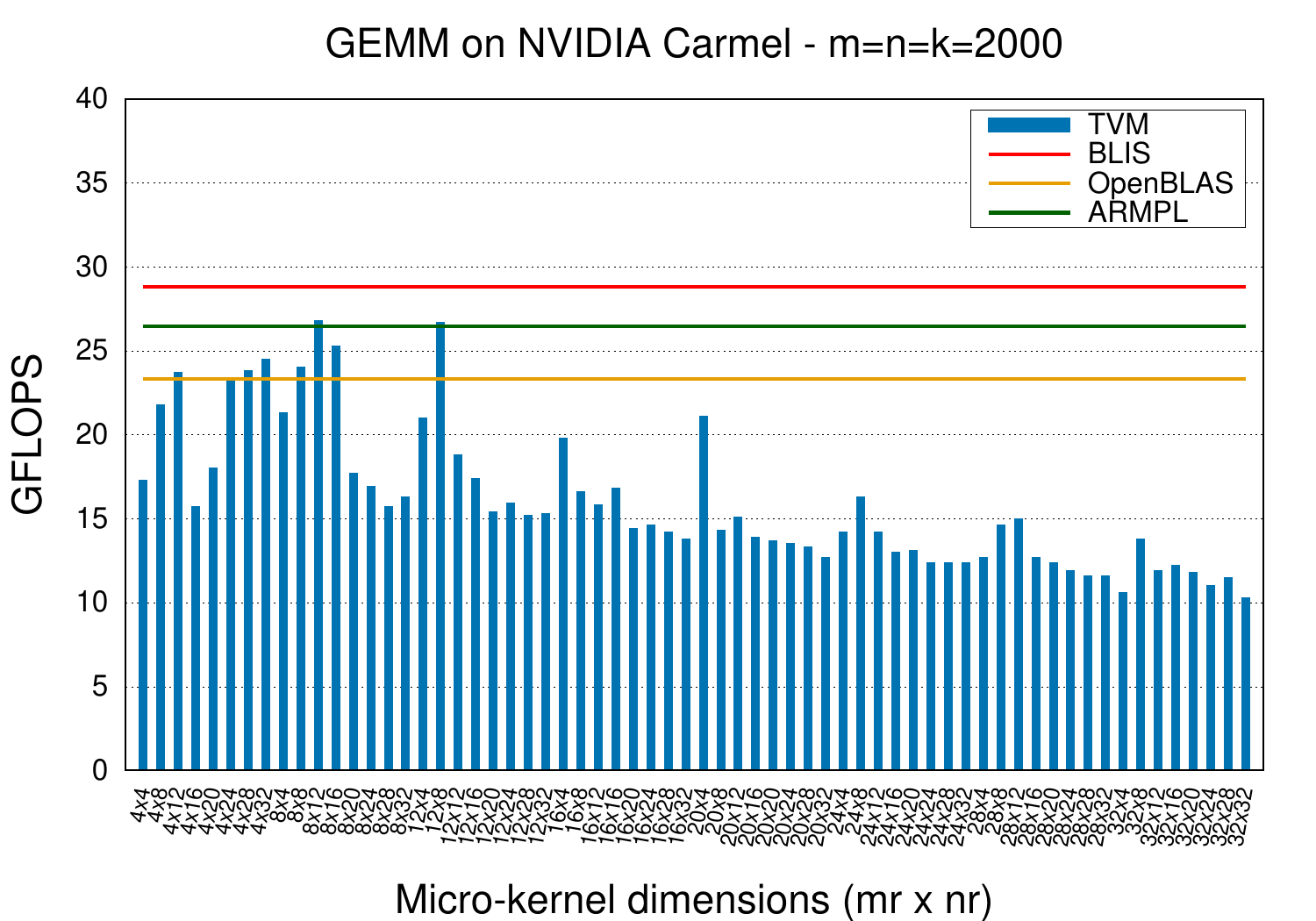}
\includegraphics[width=0.48\textwidth]{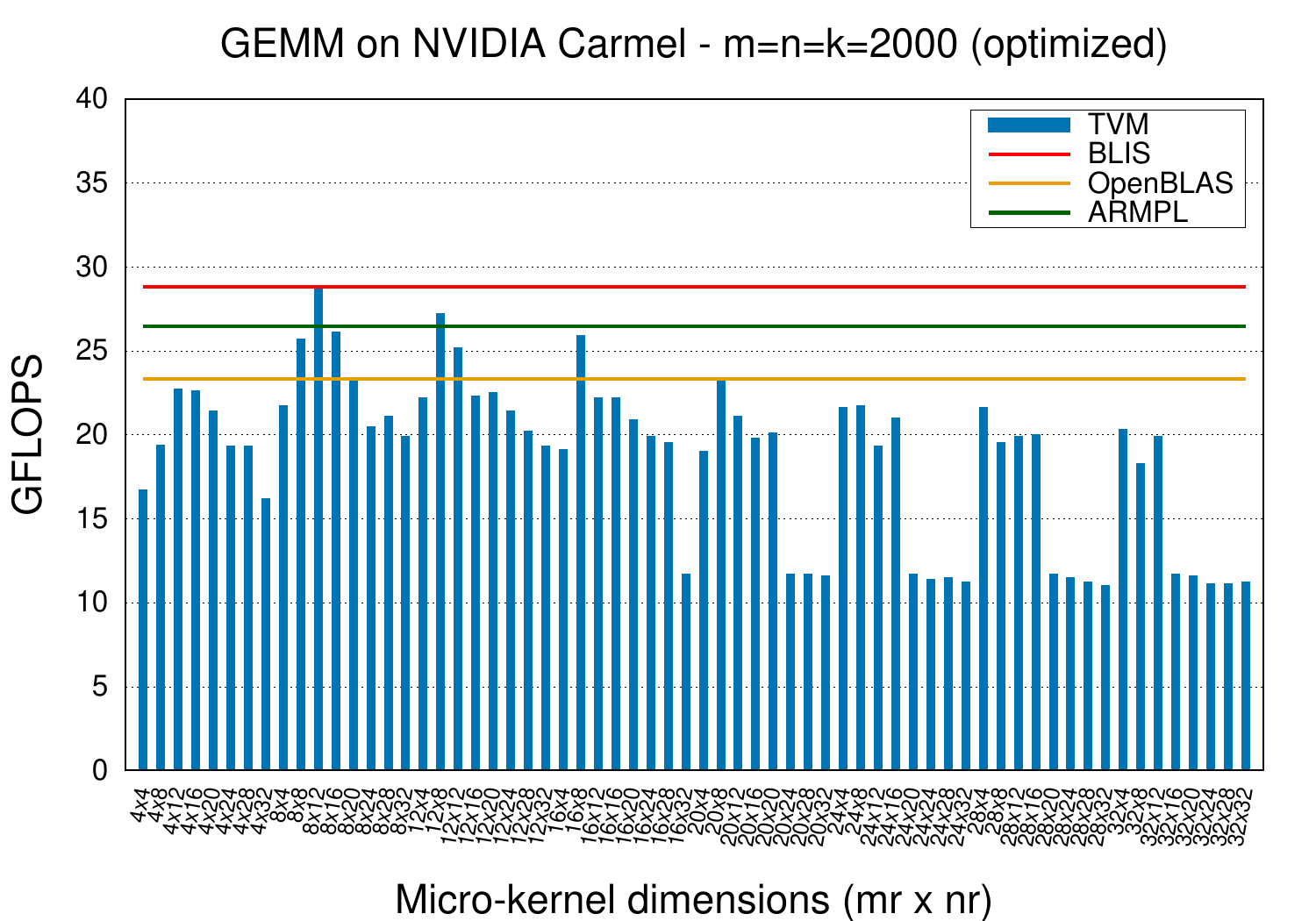}
\end{center}
\caption{Performance evaluation of the TVM generator on a single NVIDIA Carmel core for square matrices, without and with fixed lane size/prefetching (left and right, respectively).}
\label{fig:microkernel_ARM_square} 
\end{figure}

\begin{figure}[thb]
\begin{center}
          \includegraphics[width=0.48\textwidth]{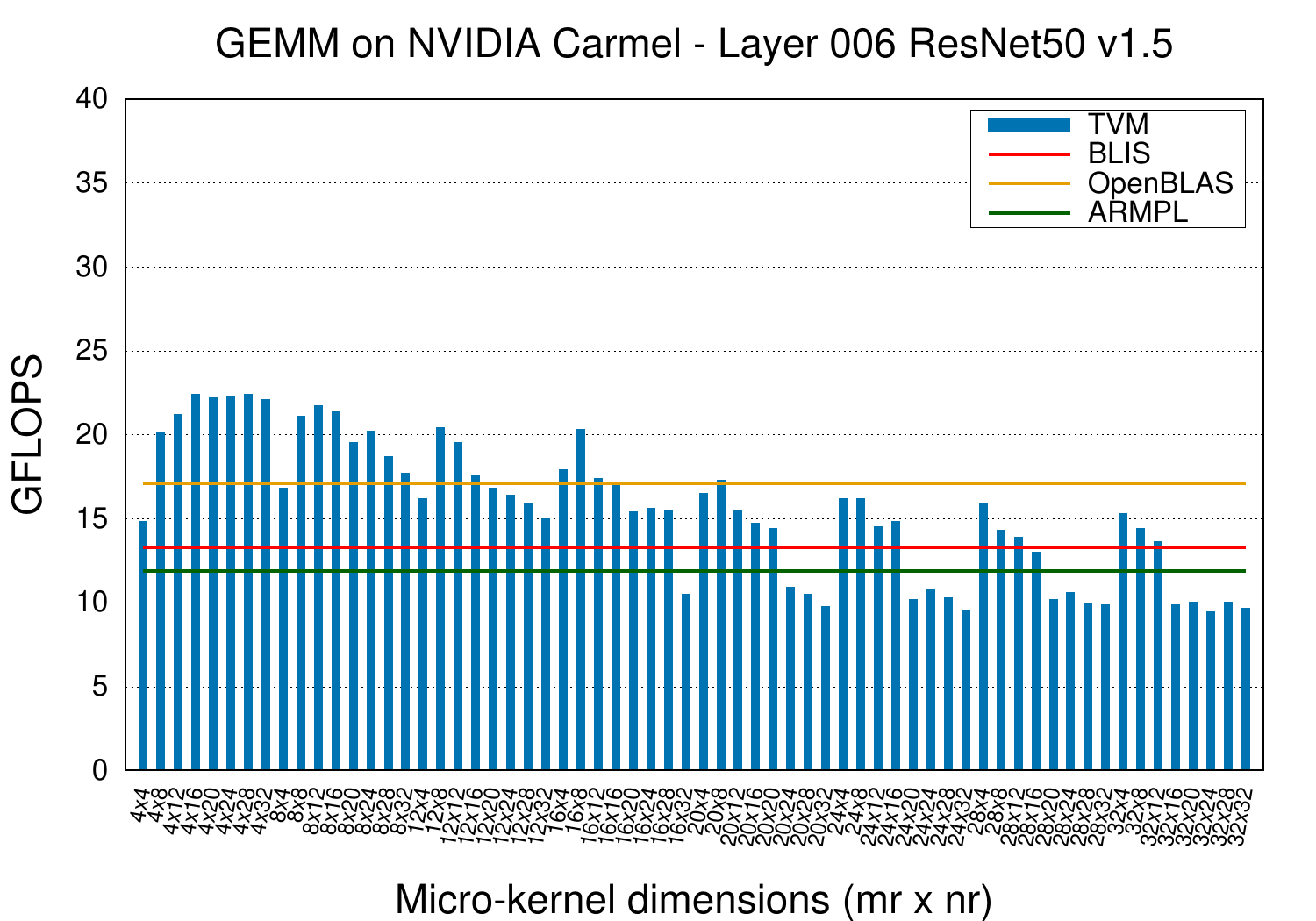}
          \includegraphics[width=0.48\textwidth]{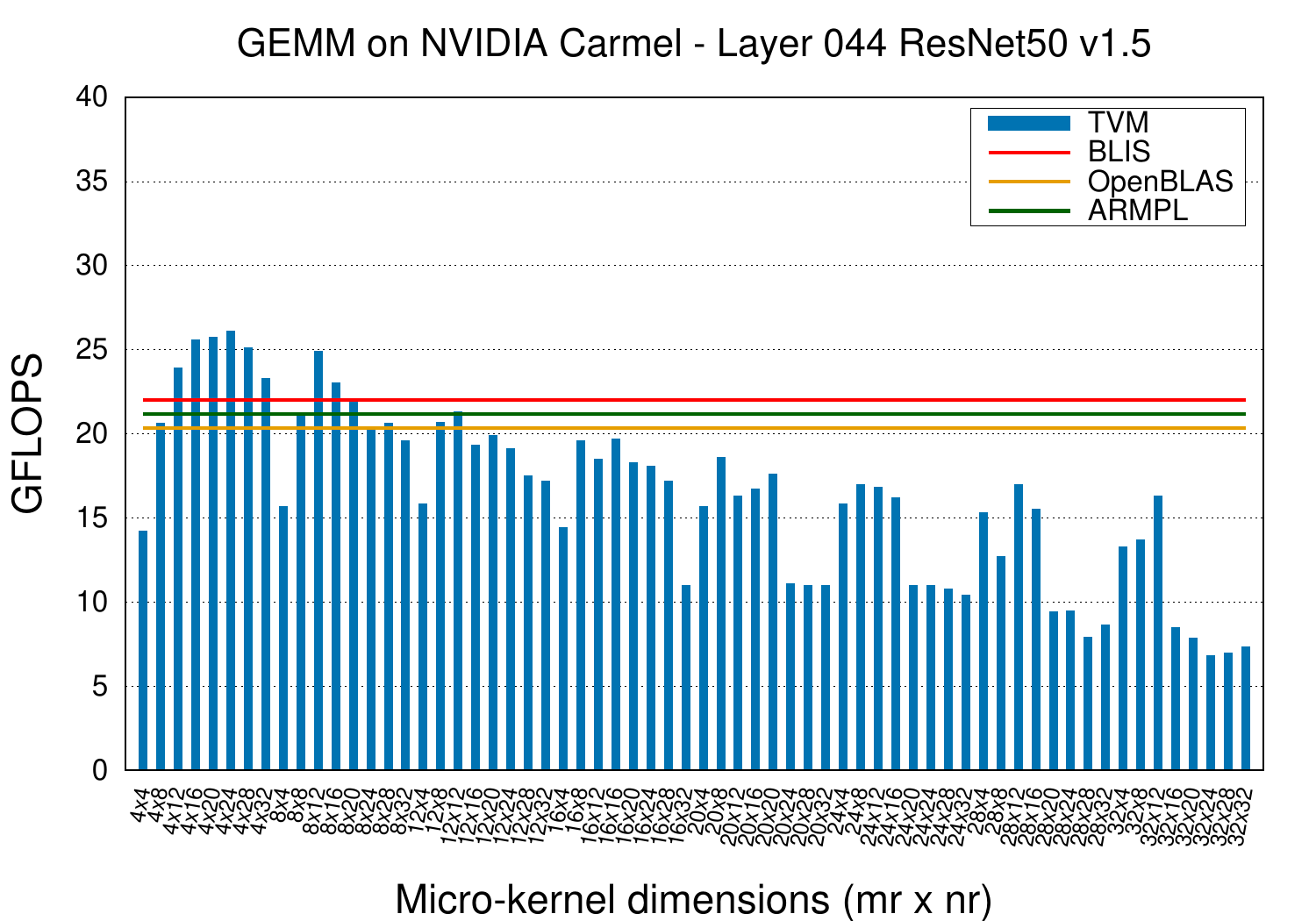}
\end{center}
\caption{Performance evaluation of the TVM generator on a single NVIDIA Carmel core for layers 2 (left) and 8 (right) of ResNet50 v1.5.}
\label{fig:microkernel_ARM_rectangular} 
\end{figure}




\subsection{In search of the best micro-kernel}

In this subsection we demonstrate the benefits of being able to 
automatically generate and subsequently
evaluate a variety of micro-kernels for a particular problem dimension. 
(In comparison,
a manual development is a costly process that requires a high 
level of expertise.)

\subsubsection{Effect of the micro-kernel}
\label{subsec:lanesize}

We unfold the analysis of the micro-kernels
by assessing the performance of the TVM-generated routines 
for the baseline algorithm
that integrate micro-kernels of
different dimensions: For a square problem with $m=n=k=$~2,000
in Figure~\ref{fig:microkernel_ARM_square},
and for the \gemm operations arising in two layers
of the ResNet model 
in Figure~\ref{fig:microkernel_ARM_rectangular}.
We clarify here that the only difference between these micro-kernels is their
dimensions. Thus, other 
{\color{black}
optimization possibilities, such as varying the loop unrolling factor, 
selecting the loop that to vectorize, etc. were not evaluated in this experiment and were kept constant across all experiments.
}
{\color{black}With respect to the two plots in
Figure~\ref{fig:microkernel_ARM_square}, 
they evaluate the impact of two low-level optimizations 
included in the TVM generator in Figure~\ref{lst:opt_GEMM_w_ukernel}:
}
\begin{itemize}
    \item \textit{Lane size:} 
    The innermost loop of \gemm is further 
    split using a factor that is an integer multiple of the number of 
          elements that fit into one vector register (e.g., 
          8 for FP16, 4 for FP32, or 2 for FP64 
          in the NVIDIA Carmel processor, for which 
          the vector registers are 128-bit wide).
    \item \textit{Prefetching:} The TVM method \texttt{\small cache\_read} is invoked to induce 
          the scheduler to include an  automatic \textit{prefetching} when possible. 
\end{itemize}
{\color{black}
The left plot in Figure~\ref{fig:microkernel_ARM_square} was
obtained by removing Part 6.4 (lines 24 to 38) in
the TVM generator in  Figure~\ref{lst:opt_GEMM_w_ukernel}, 
while the right plot was obtained with these lines included.
In both cases, the performance results correspond to sequential
executions.
}

As a result from these additional optimizations,
the performance of the TVM routine
displayed in the right plot of Figure~\ref{fig:microkernel_ARM_square},
when setting the micro-kernel dimension to 
$m_r \times n_r = 8\times 12$, 
improves from 16 to 18.7 GFLOPS with respect 
to its counterpart without these low-level optimizations. 
Therefore, from now on we will only report results for the TVM-generated routines
with these two optimizations in place.

The right plot in Figure~\ref{fig:microkernel_ARM_square}
demonstrates that a careful selection of the 
micro-kernel is critical to attain
high performance. Concretely,
the performance achieved by using micro-kernels of different dimensions 
varies between
\textcolor{black}{10.3} GFLOPS (worst case, with $m_r \times n_r = 32\times 32$) and \textcolor{black}{26.8} GFLOPS
(best case, with $m_r \times n_r = 8 \times 12$). %
The two performance plots in Figure~\ref{fig:microkernel_ARM_rectangular} 
also contribute to show that the best
micro-kernel is largely dependent on the problem dimension.
Specifically, the highest GFLOPS rates are observed for 
the micro-kernels $m_r \times n_r=4 \times 16$ and
$4 \times 28$ for layer 2, 
compared with the micro-kernel $m_r \times n_r = 4 \times 24$ for layer 8.

A direct comparison between the realizations of \gemm in ``hand-encoded'' libraries 
and the TVM routine 
\textit{using the best micro-kernel} for each problem
case
reveals that, for square matrices, the TVM routine delivers GFLOPS 
rates that are similar to those attained with BLIS
(\textcolor{black}{28.9} GFLOPS for the former 
versus
\textcolor{black}{28.8} GFLOPS for latter), and superior 
with respect to OpenBLAS (\textcolor{black}{23.3} GFLOPS) 
and ARMPL (\textcolor{black}{26.4} GFLOPS).
The scenario is different for the ResNet50 problems: 
For layer 2, the TVM routine (with the best micro-kernel)
delivers 
\textcolor{black}{22.4} GFLOPS versus \textcolor{black}{13.3} GFLOPS for BLIS,
17.1 GFLOPS for OpenBLAS, and 11.9 for ARMPL.
In addition, for layer 8, the TVM routine (with the best micro-kernel) achieves 
\textcolor{black}{26.1} GFLOPS versus \textcolor{black}{22.0} GFLOPS for BLIS (best library-based option).
From this point, we will always report the results for the TVM routine with
the best micro-kernel for each 
problem dimension. 

\begin{figure}[thb]
\begin{center}
          \includegraphics[width=1\textwidth]{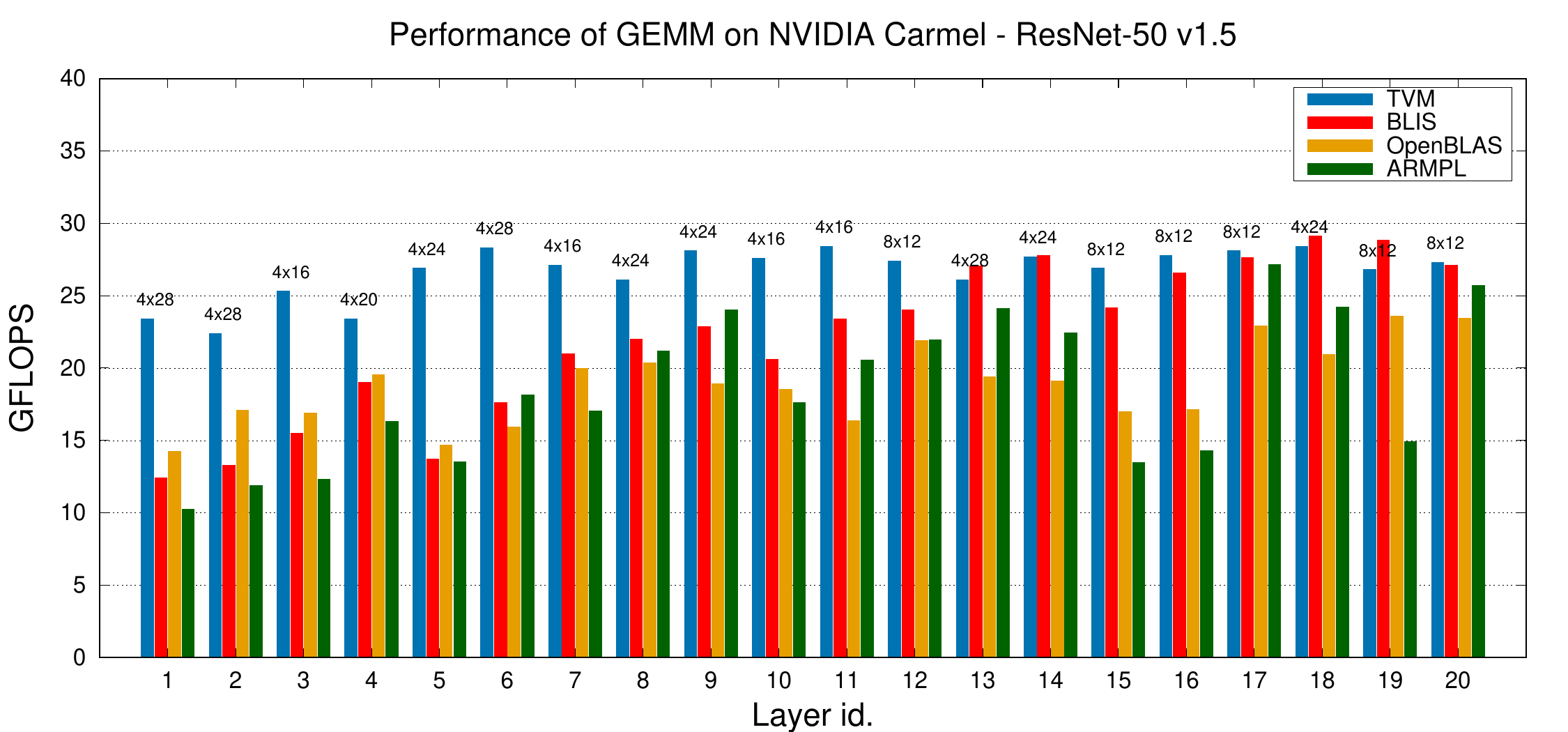}
\end{center}
\caption{Performance evaluation of the TVM generator on a single NVIDIA Carmel core for ResNet50 v1.5. The number on top of each blue bar represents the dimension $m_r \times n_r$ of the best
micro-kernel for that layer.}
\label{fig:microkernel_ARM_ResNet} 
\end{figure}

The evaluation of the \gemm operations associated with all 
the layers in the ResNet50 model 
shows a variety of scenarios. Concretely, 
Figure~\ref{fig:microkernel_ARM_ResNet} 
illustrates that the TVM routine outperforms the BLIS realization by a large margin
for layer~1--12 and by a visible difference for layers~~15--17 
(40 cases out of the total 53 convolution layers in the model, see Table~\ref{table:resnet50}); 
it is competitive for layer~14 and~20 (3 cases); and it is suboptimal
for layer~13, 18, 19 (10 cases). 
Compared with OpenBLAS and ARMPL, the TVM routine is consistently better.
An analysis of the results taking into account the operands' dimensions 
shows that the TVM routine delivers higher performance for 
``rectangular'' cases, with
$m$ in the range 100,352--1,605,632, and it is competitive when $m$=25,088. 
In contrast, BLIS is better choice for ``square'' problems,
with $m$ in the range of 6,000.%
\footnote{As a side note, 
the actual processing cost of the Resnet50 v1.5 model is concentrated in those cases
where $m$ is in the range 100,352--1,605,632 
(47.8\% of the total time), followed by $m$=25,088 (35.6\% of the total time).
In terms of absolute cost, this implies that
the execution of all layers employing the TVM
routine would require 39.1~s compared with 48.0~s when 
using BLIS (and higher for OpenBLAS and ARMPL).}




\subsubsection{Why is TVM better?}
The superiority of the TVM routine is rooted in the fact that, by (automatically)
generating the micro-kernels of different dimensions,
we can easily explore the space and select the one that is
better suited to a particular problem dimension. 
This is illustrated in
Figure~\ref{fig:microkernel_ARM_ResNet}, which reports the best micro-kernel for each
problem dimension/DNN layer.
Compared with that, 
BLIS, OpenBLAS and ARMPL each integrate a single, manually-encoded micro-kernel, which is therefore the only option
for any problem dimension.
The reason for this limitation of the libraries is that 
manually implementing different micro-kernels 
is a time-consuming task, requiring significant experience in high performance computing, computer architecture, and assembly coding.
In addition, the logic of selecting the appropriate micro-kernel dimension
based on problem dimensions is usually not supported in 
commercial or academic libraries.

%
Delving further into the matter,
the theoretical reasons behind this behavior can be explained using 
the analytical model in~\cite{BLIS4}. 
According to that, for problems with a reduced dimension $k$,
which in practice limits the effective value for the cache
parameter $k_c$,  the micro-panel of $B_c$ that is stored in the L1 cache  
($B_r$, of dimensions $k_c \times n_r$) does not attain the optimal 
occupation of that level, explaining the performance
penalty. 

Let us illustrate the problem for layer 2 and 8 of
ResNet v1.5, whose experimental performance for different micro-kernel dimensions 
was exposed in Figure~\ref{fig:microkernel_ARM_rectangular}.
%
We complement that figure with the values in Table~\ref{tab:L1_occupation_layers2_8}, 
illustrating the cache parameters 
$m_c, n_c, k_c$ and the theoretical occupation of the L1 cache by $B_r$, for 
different micro-kernel dimensions, 
determined using the analytical model in~\cite{BLIS4}\footnote{In the table, we limit the 
study to those values of $m_r$ and $n_r$ that do not
cause register spilling, as this effect would yield a performance penalty
beyond that introduced by the negative effect of L1 under-occupation.}.
When executed on the Carmel processor, the analytical model 
indicates that, for the $m_r \times n_r = 8 \times 12$ micro-kernel that
is integrated BLIS for that particular architecture,
the micro-panel $B_r$ targeting the L1 cache only occupies
4.69\% of the L1 cache for layer 2; and 9.38\% for layer 8.
In contrast, this micro-kernel should have occupied 
up to 50\% with $B_r$ for both layers, reserving the rest of the L1 cache 
for entries from the $A,C$ operands.

The only way to address this under-utilization of the L1 cache is by 
increasing $n_r$, and hence developing a new micro-kernel. 
Let us support this observation with a specific example.
Consider a micro-kernel of dimension
$m_r \times n_r = 4\times 28$. 
In this case,  the occupancy of the L1 cache by $B_r$ grows to 10.90\% 
for layer 2 and up to 21.90\% for layer 8,
which is  clearly superior to that observed
for the (BLIS) $8 \times 12$ micro-kernel.
In summary, there is a clear theoretical benefit from
adopting an $m_r \times n_r=4 \times 28$ micro-kernel
for these particular layers, which is conformal with 
the experimental advantage that was reported in
Figure~\ref{fig:microkernel_ARM_ResNet}.

Figure~\ref{fig:microkernel_ARM_Cache_usage} reports the
utilization rates of the L1 and L2 cache levels by $B_r$ and $A_c$,
respectively, and all the layers. 
The results show that, especially for the L1 cache, the 
occupation compared with BLIS explains the overall differences in performance between the BLIS micro-kernel and the
optimal alternative automatically generated with TVM reported in Figure~\ref{fig:microkernel_ARM_ResNet}.  
For reference, the figure includes the theoretical maximum occupation rate for each cache memory (black line) for the corresponding operands, dictated by the analytical model. 

To wrap up this discussion, the under-utilization of the L1 and L2 cache levels
resulting from the small values of $k$ and $m$ in non-square problems forces the
developer to trade it off with larger register block dimensions 
($m_r$ and/or $n_r$), and 
hence with the development a family of micro-kernels that, in practice,
are invoked intelligently depending on the problem dimensions.
Using a tool like TVM to automatically generate micro-kernels, together
with an analytical model for cache and register blocking parameters,
alleviates this programmability burden not only to obtain 
dimension-agnostic performance optimization on a single architecture, 
but also across different architectures.

\begin{table}[]
\footnotesize
\caption{Detailed L1 cache occupation for layers 2 and 8 of ResNet50 v1.5 for different micro-kernel dimensions.}
\label{tab:L1_occupation_layers2_8}
\begin{tabular}{ccccccccccccccccc}
\toprule
Layer & $m_c$   & $n_c$  & $k_c$  & $m_r$ & $n_r$ & L1   & Max &&
Layer & $m_c$   & $n_c$  & $k_c$  & $m_r$ & $n_r$ & L1   & Max \\
type id. &         &        &        &       &       & (\%) & (\%) &&
type id. &         &        &        &       &       & (\%) & (\%) 
\\ \cline{1-8} \cline{10-17}
2  & 7,168 & 64  & 64  & 4  & 4  & ~~1.56 & 25 && 
8  & 3,584 & 256 & 128 & 4  & 4  & ~~3.12 & 25 \\
2  & 6,656 & 64  & 64  & 4  & 8  & ~~3.12 & 50 &&
8  & 3,584 & 256 & 128 & 4  & 8  & ~~6.25 & 50 \\
2  & 6,144 & 64  & 64  & 4  & 12 & ~~4.69 & 50 &&
8  & 3,328 & 275 & 128 & 4  & 12 & ~~9.38 & 50 \\
2  & 6,144 & 64  & 64  & 4  & 16 & ~~6.25 & 50 &&
8  & 3,328 & 275 & 128 & 4  & 16 & 12.50  & 50\\
2  & 5,632 & 64  & 64  & 4  & 20 & ~~7.81 & 50 &&
8  & 3,072 & 298 & 128 & 4  & 20 & 15.60  & 50 \\
2  & 5,120 & 64  & 64  & 4  & 24 & ~~9.38 & 50 &&
8  & 3,072 & 298 & 128 & 4  & 24 & 18.80  & 50\\
2  & 5,120 & 64  & 64  & 4  & 28 & 10.90  & 50&&
8  & 3,072 & 298 & 128 & 4  & 28 & 21.90  & 50\\
2  & 7,168 & 64  & 64  & 8  & 4  & ~~1.56 & 25 &&
8  & 3,584 & 256 & 128 & 8  & 4  & ~~3.12 & 25 \\
2  & 6,656 & 64  & 64  & 8  & 8  & ~~3.12 & 25 &&
8  & 3,584 & 256 & 128 & 8  & 8  & ~~6.25 & 25 \\
2  & 6,144 & 64  & 64  & 8  & 12 & ~~4.69 & 50 &&
8  & 3,328 & 275 & 128 & 8  & 12 & ~~9.38 & 50 \\
2  & 7,168 & 64  & 64  & 12 & 4  & ~~1.56 & 25 &&
8  & 3,584 & 256 & 128 & 12 & 4  & ~~3.12 & 25 \\
2  & 6,656 & 64  & 64  & 12 & 8  & ~~3.12 & 25 &&
8  & 3,584 & 256 & 128 & 12 & 8  & ~~6.25 & 25 \\
2  & 7,168 & 64  & 64  & 16 & 4  & ~~1.56 & 25 &&
8  & 3,584 & 256 & 128 & 16 & 4  & ~~3.12 & 25 \\
2  & 7,168 & 64  & 64  & 20 & 4  & ~~1.56 & 25 &&
8  & 3,584 & 256 & 128 & 20 & 4  & ~~3.12 & 25 \\
2  & 7,168 & 64  & 64  & 24 & 4  & ~~1.56 & 25 &&
8  & 3,584 & 256 & 128 & 24 & 4  & ~~3.12 & 25 \\
2  & 7,168 & 64  & 64  & 28 & 4  & ~~1.56 & 25 &&
8  & 3,584 & 256 & 128 & 28 & 4  & ~~3.12 & 25 \\ \bottomrule
\end{tabular}
\end{table}

\begin{figure}[thb]
\begin{center}
          \includegraphics[width=0.49\textwidth]{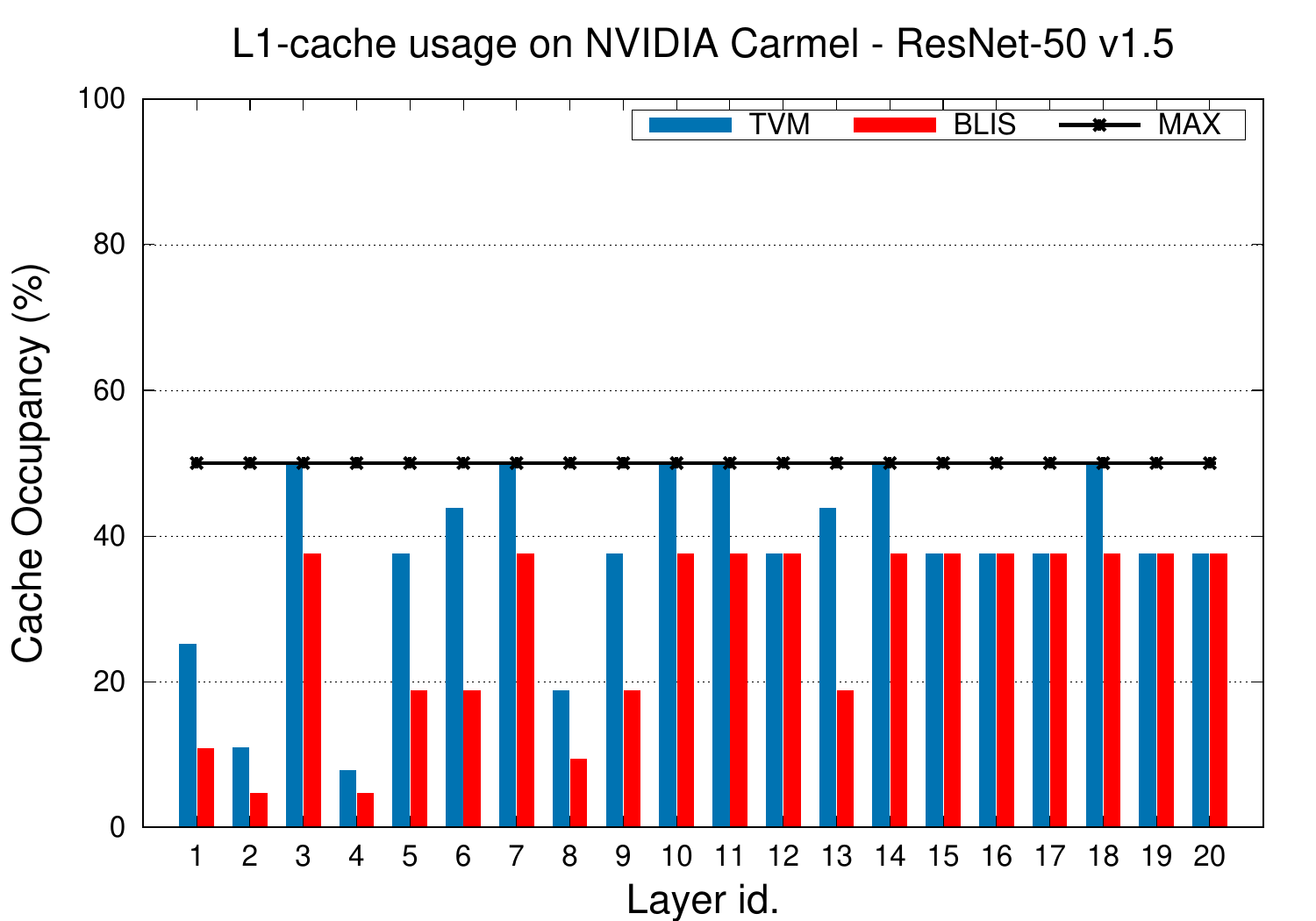}
          \includegraphics[width=0.49\textwidth]{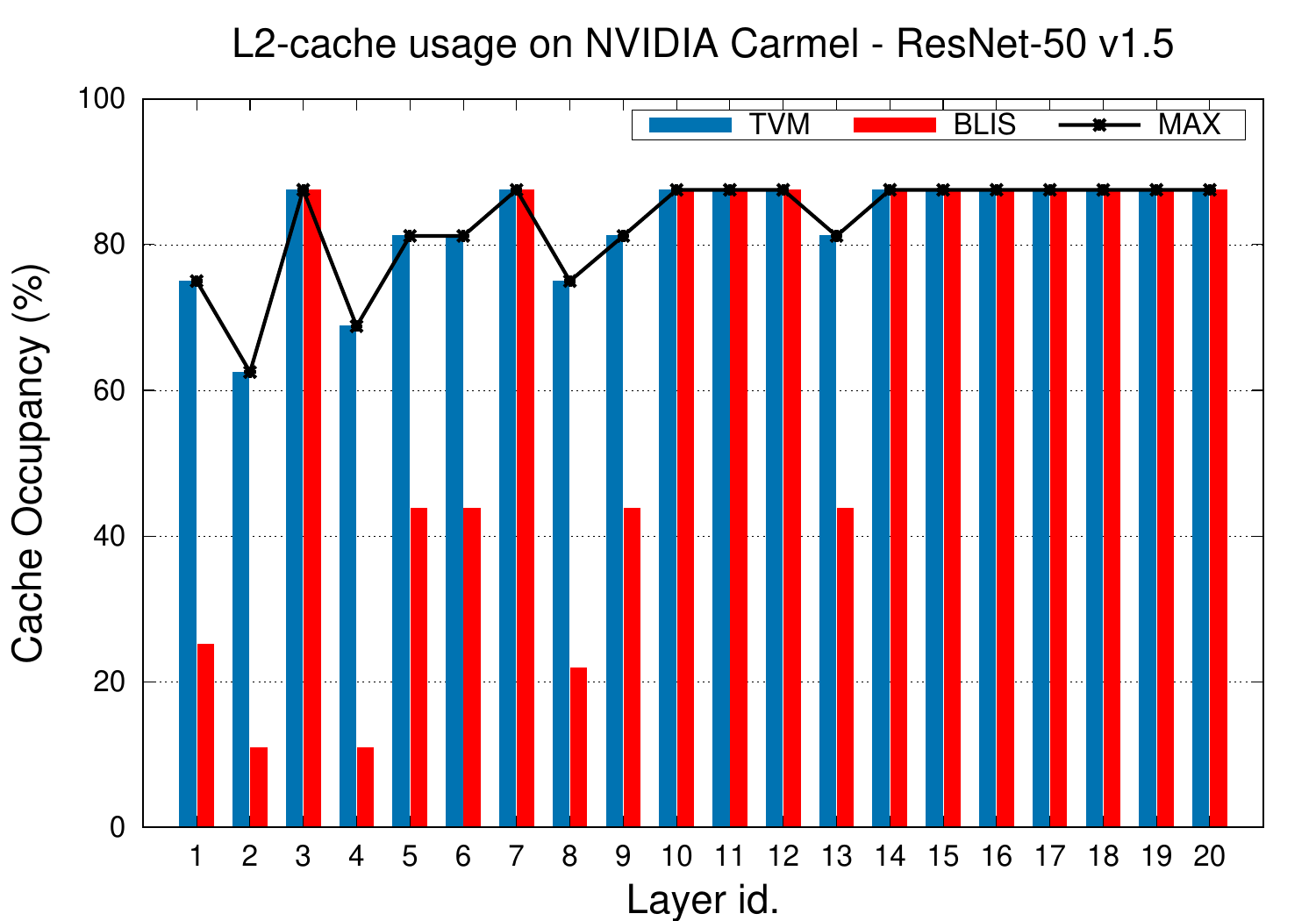}
\end{center}
\caption{Occupation for the L1 and L2 caches (left and right, resp.) on the Carmel platform for the best TVM-generated micro-kernel compared with the BLIS micro-kernel.} 
\label{fig:microkernel_ARM_Cache_usage} 
\end{figure}


\subsubsection{Optimization of the micro-kernel}

To close this discussion on the impact of the micro-kernel, 
we note that, 
when searching for the optimal dimension of the micro-kernel, 
the exploratory space is basically 
bi-dimensional, as we only need to select the values for $m_r,n_r$. Now, 
the optimal value for at least one of 
these two parameters is tied to the architecture lane size and,
in addition, the hardware imposes strict limits on the number of vector registers
that can be used from inside the micro-kernel, which in turn constrains the practical values for $m_r,n_r$.
As a consequence, 
selecting the best option for the TVM-generated routines
only requires a few dozens of experiments, and the whole
process can be also automatized, providing significant benefits in terms of reduced 
programming and optimization efforts for the library developer.

In the next subsections \textit{we examine the programming benefits
of the TVM-based approach,
from the viewpoints of performance, maintainability and portability,}
by exposing how TVM allows us to easily generate routines for different data types,
explore distinct packing schemes, 
evaluate alternative parallelization options and, finally, 
instantiate the full family of matrix
multiplication algorithms.
Finally, we close the experimental section with an evaluation on AMD
and Intel architectures.

{\color{black}
\subsection{Maintainability: Generating codes for different data types}


Generating a routine for a specific data type with TVM only requires adjusting the \texttt{dtype} argument
in the TVM generator, and modifying accordingly the lane size (see subsection~\ref{subsec:lanesize}). 
Compared with this, producing manually a GotoBLAS-2 like routine, 
for a particular data type, requires a careful re-design of the micro-kernel,
usually in assembly, as well as the adaptation of the packing functions.

In Figure~\ref{fig:Gflops_datatypes} we report the performance of the  routines generated with TVM
for five data types: 
FP16 (IEEE 16-bit floating point), 
FP32 (IEEE 32-bit floating point), 
FP64 (IEEE 64-bit floating point), 
INT16 (integer 16-bit), and 
INT32 (integer 32-bit).
This figure shows acceleration factors which are
conformal with the use of
other precision formats. 

\begin{figure}[tbh]
\begin{center}
        \includegraphics[width=1\textwidth]{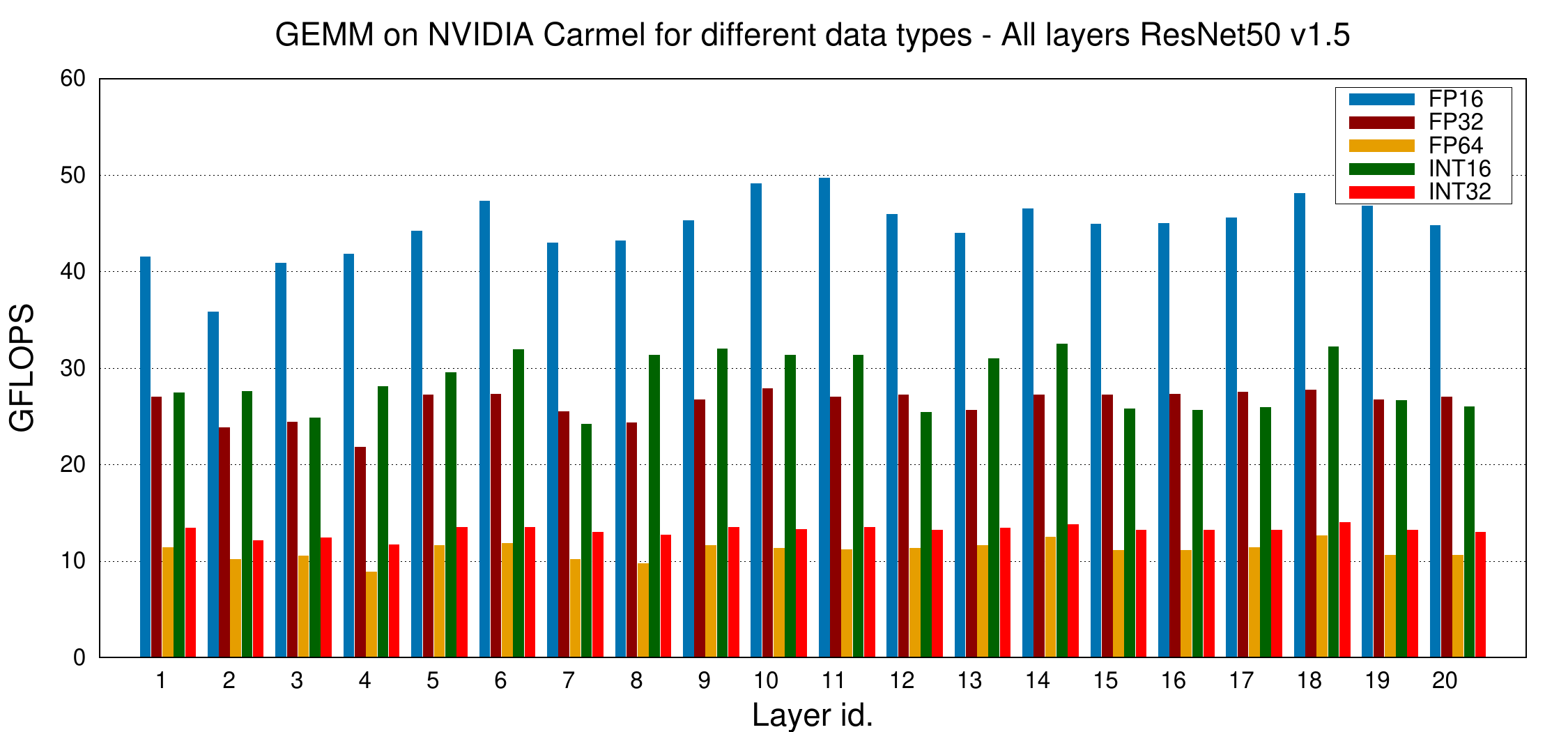}
\end{center}
\caption{Performance of the TVM generator for distinct data types 
on a single NVIDIA Carmel core
for ResNet-50 v1.5.
}
\label{fig:Gflops_datatypes} 
\end{figure}


\subsection{Performance: Packing costs}

{\color{black}For some problem dimensions, 
it may be beneficial to skip the packing of any of the matrix operands, or both of them, into 
the buffers $A_c,B_c$. 
As described in subsection~\ref{subsec:packing},
eliminating
the packing scheme is straight-forward with TVM.
In contrast, introducing this modification into
a conventional GotoBLAS2-like routine implies 
rewriting the micro-kernel, as this piece of code
assumes that the matrix operands
are disposed/packed into the buffers in a certain manner;
see Figure~\ref{fig:blis_packing}. Given that the micro-kernel
is in general encoded in assembly, this is a non-trivial task.

Figure~\ref{fig:Gflops_packing} 
evaluates the packing possibilities using TVM to generate modified versions
of the baseline algorithm. 
As shown in the top chart, for small square matrices, 
the cost of the re-arranging the data due to packing is not compensated with a ``sufficient'' 
acceleration of the
micro-kernel. For the problems associated with the convolutional layers in ResNet-50 v1.5 model, 
the dimensions are large enough and this effect is not present. 
Nonetheless, the irregular sizes of these operands dictate that, for 
layers~1, 2, 5, 7 and~18, the performance of the baseline algorithm 
generated with TVM and a variant which packs only one of the matrix operands are close. 
}

\begin{figure}[tbh]
\begin{center}
        \includegraphics[width=1\textwidth]{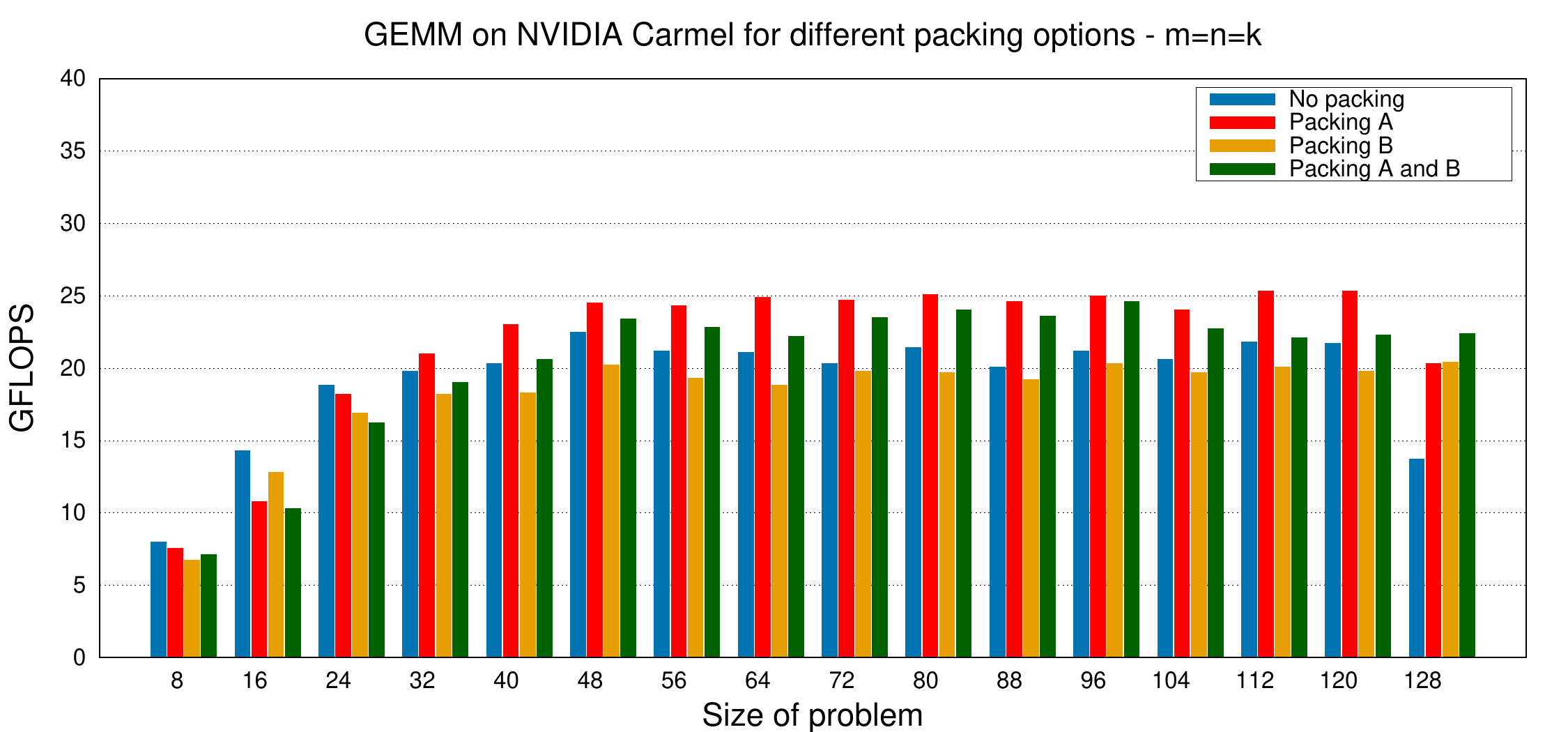}
        \includegraphics[width=1\textwidth]{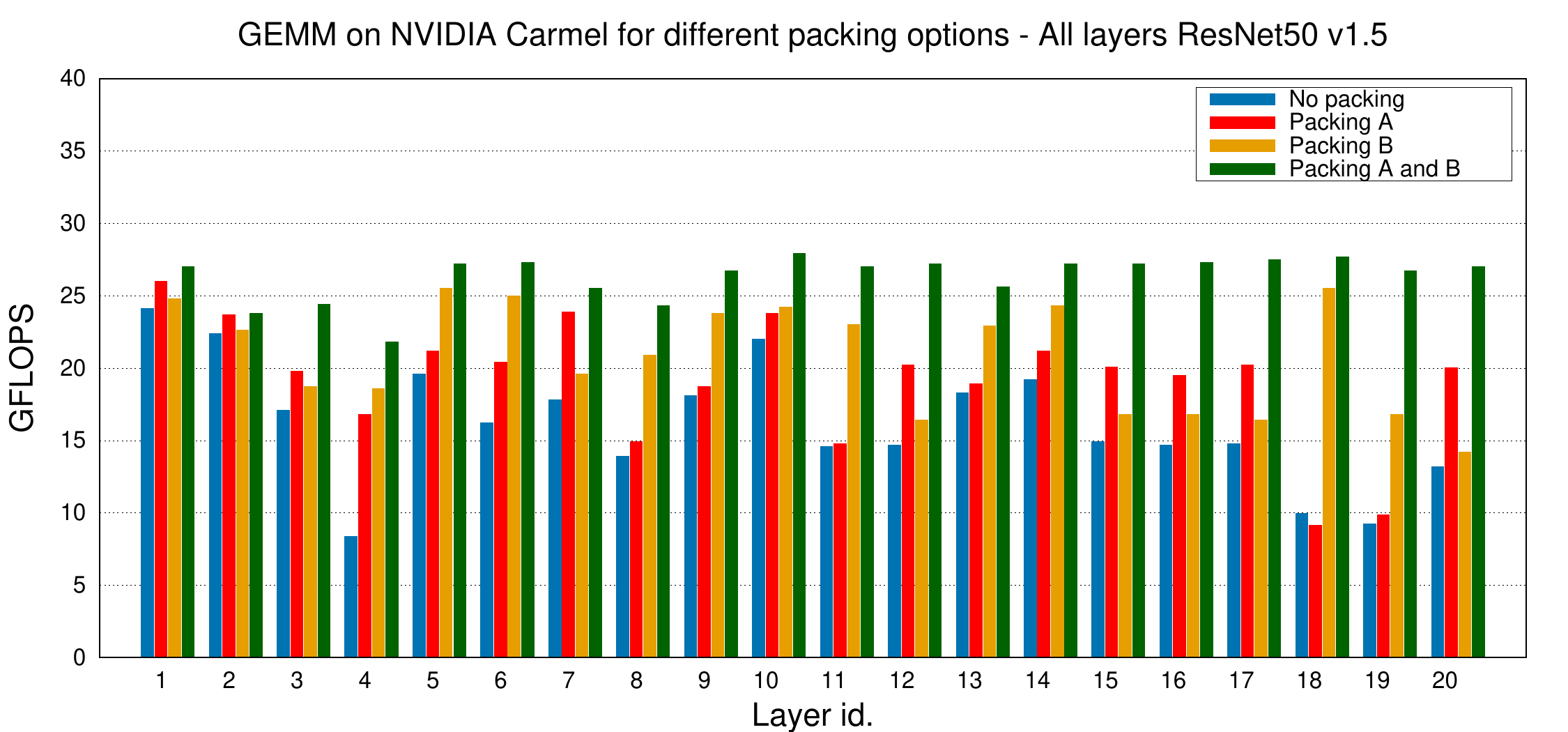}
\end{center}
\caption{Performance of the TVM generator for four distinct packing configurations 
         using a single NVIDIA Carmel core for small-square matrices and ResNet-50 v1.5 (top and bottom, respectively).}
\label{fig:Gflops_packing} 
\end{figure}


\subsection{Performance: Parallelization options}

The TVM generator can be leveraged to assess 
distinct options to orchestrate a multi-threaded execution on a multicore processor.
In Figure~\ref{fig:Gflops_parallel_versions}, we target the 8~cores 
in the NVIDIA Carmel processor, evaluating four parallelization options 
for the baseline algorithm (see Figure~\ref{fig:blis_family_Cresident}, left)
that differ in the loop which is parallelized:
$j_c,i_c,j_r$, or $i_r$. 
(Loop $p_c$ cannot be parallelized as this would yield a race condition.)
This shows  that the best choice varies 
slightly between two of the options, depending on the problem dimensions.
However, investigating the best parallelization option adds a dimension to the complexity 
to the optimization effort that is out-of-scope for this work.

\begin{figure}[!t]
     \centering
           \includegraphics[width=1\textwidth]{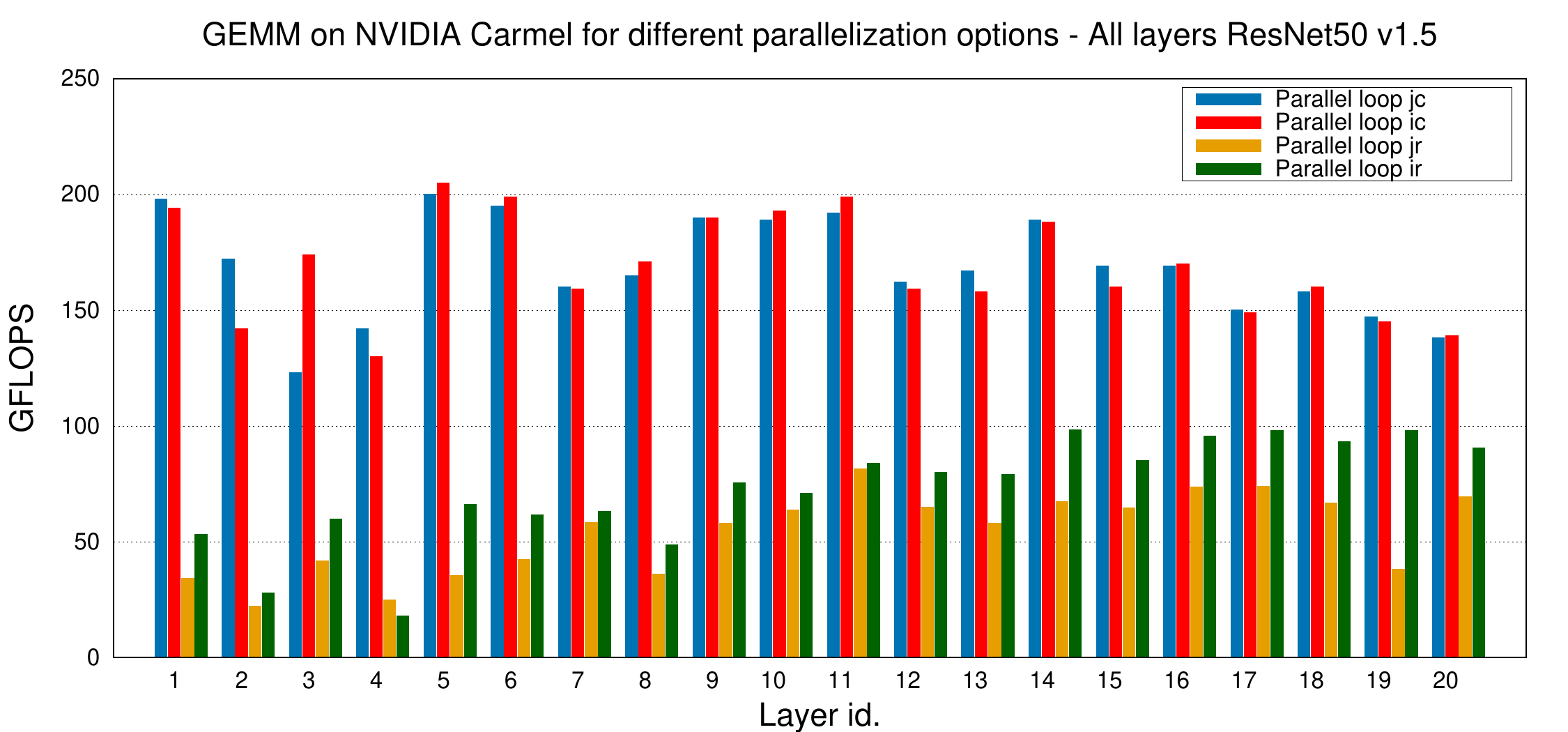}
\caption{Performance of the TVM generator for four distinct loop 
         parallelization options 
         on 8 NVIDIA Carmel cores for ResNet-50 v1.5.}
        \label{fig:Gflops_parallel_versions}
\end{figure}


At this point 
we recognize that an analysis of the parallelization options for a conventional
GotoBLAS-2-like routine is also straight-forward.
Furthermore, 
in some cases, it may be more convenient to parallelize multiple loops to expose
sufficient thread-level parallelism%
~\cite{BLIS2}. 
Unfortunately, when instructed to parallelize two or more loops, currently 
TVM extracts parallelism from the outermost loop only.






\subsection{Performance, maintainability and portability: The complete family of algorithms}

In Section~\ref{sec:otherfamily}, we argued that
producing code with TVM for other blocked algorithmic variants of the \gemm family 
only requires small changes into the generator routines.
In this subsection we analyze the practical impact of this on 
the sequential and parallel performance.

\begin{figure}[t!]
\begin{center}
         \includegraphics[width=1\textwidth]{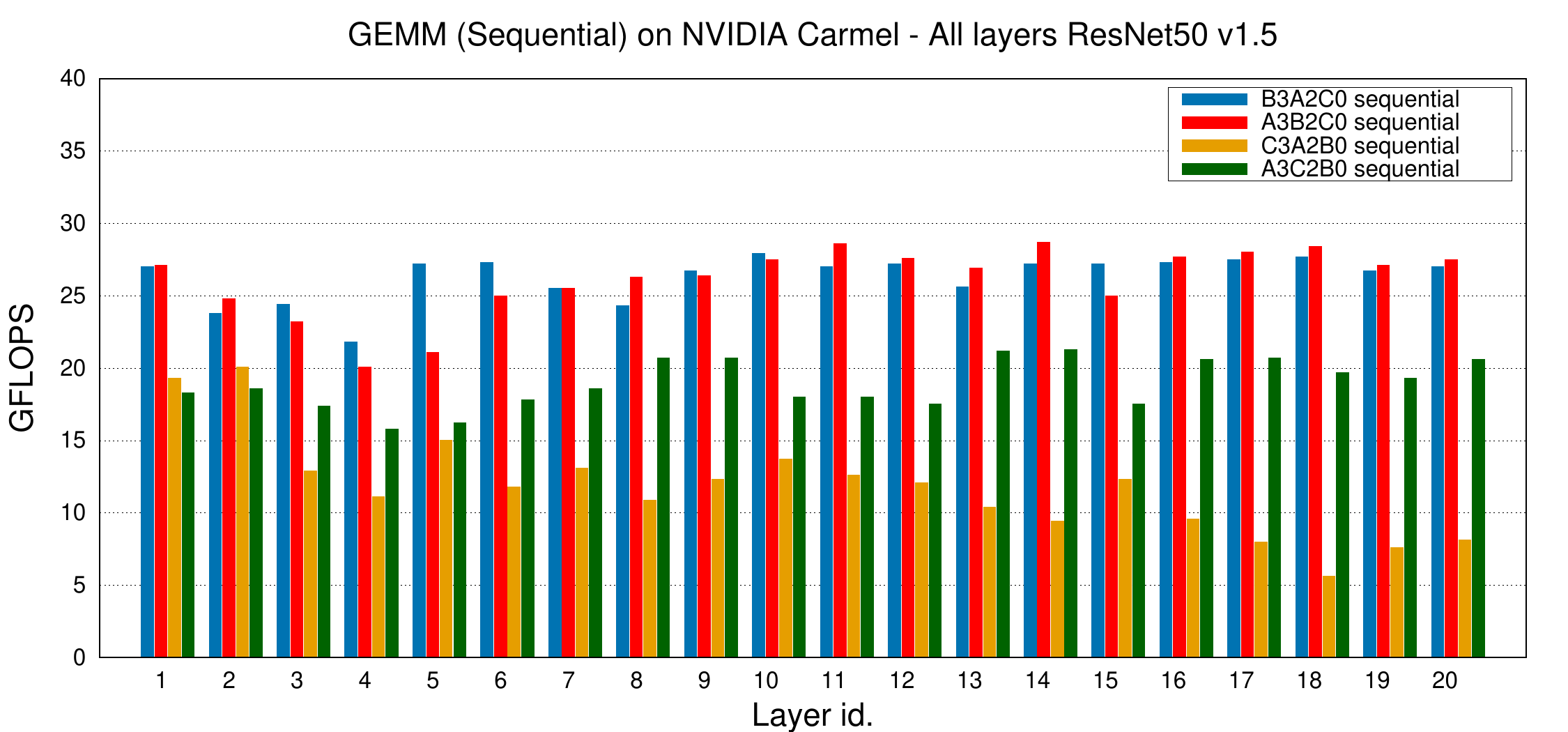}
         \includegraphics[width=1\textwidth]{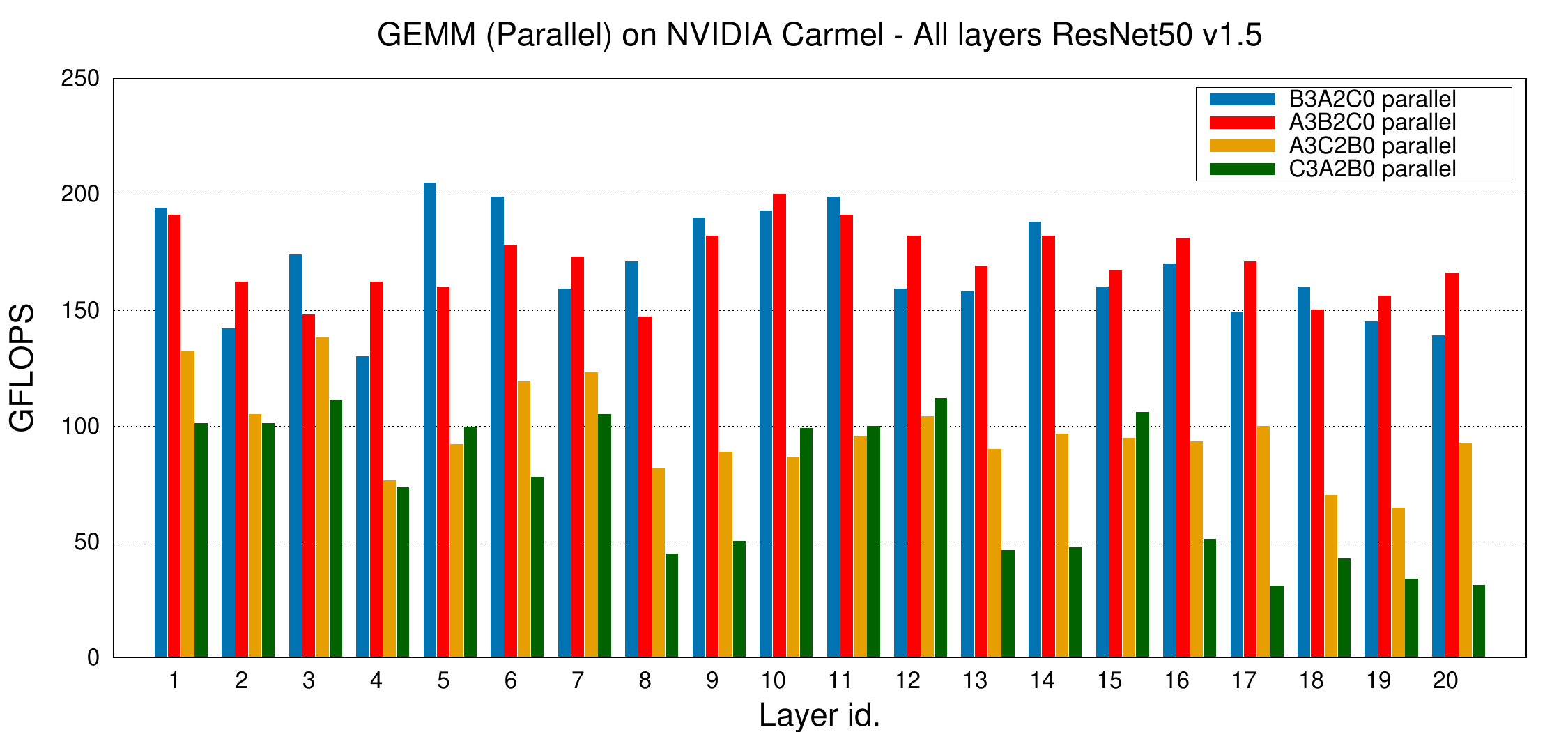}
\end{center}
\caption{Performance of the TVM generators for four variants of the 
         \gemm family of algorithms on a single core and 8 cores 
         of the NVIDIA Carmel processor (top and bottom, respectively),
         for ResNet-50 v1.5. Note the different scales of the $y$ axis in the two plots.}
\label{fig:all_parallel} 
\end{figure}
}



Figure~\ref{fig:all_parallel} shows a clear superiority of the variants that maintain $C$ 
in the processor registers over their counterparts that operate with $B$ in the registers.
(We omit the variants with \Aresident because they present a symmetric role with the respect to the
variants with \Bresident.) 
The reason is that, in the variants with \Cresident (i.e., B3A2C0 and A3B2C0), 
the elements of $C$ are housed in the processor registers during 
the full execution of the micro-kernel loop and, therefore, 
there are no writes to memory  as part of its innermost loop (\texttt{L6}) of the algorithm.
In contrast, the two other variants, A3C2B0 and C3A2B0, 
integrate a micro-kernel that, 
at each iteration of the innermost loop (\texttt{L6}),
performs several writes on $C$
while this operand resides in a certain level of the cache hierarchy. 
This behavior lies at the core of the algorithms/variants, and is preserved by TVM 
which simply follows the programmer's directives. 
With respect to the results, we observe small differences in performance between the
two versions that maintain $C$ in the processor registers, 
in favor of either one or another depending on the specific layer.

{\color{black}
\subsection{Memory performance: footprint}

We next investigate the 
memory requirements of the automatically-generated codes for \gemm.
{\color{black}
Table~\ref{table:memoryfootprint} reports the size 
for a bare \gemm test driver routine statically 
linked with each library (column labeled as ``\gemm''). 
}
The memory allocation for the matrices and the necessary
packing buffers are not included, as their space requirements should be similar in all cases.
Furthermore, the test driver is the same in all cases to allow a fair comparison.

\begin{table}
\centering
\caption{Size (in bytes) of the \gemm realization for each solution.}
\label{table:memoryfootprint}
\begin{tabular}{lr}
\toprule
 Solution & \gemm  \\ \midrule
ARMPL      &  29,145,688\\
BLIS       &  1,350,456 \\
OpenBLAS   &  90,944 \\
TVM        &  532,976 \\
 \bottomrule
\end{tabular}
\end{table}

%
The lowest memory footprint for the \gemm executable is offered by OpenBLAS with close to 89~KiB, followed by TVM with 520~KiB. BLIS and ARMPL need a total amount of 1.3~MiB and 27.8~MiB, respectively.
}


\subsection{Portability: Experiments in other architectures}

We close the experimental section by demonstrating 
the performance portability of the automatic-generation approach for \gemm using 
a pair of AMD and Intel processor architectures. 
For this purpose, we compare the (sequential) 
routine generated by TVM, combined with
the best micro-kernel, against that of the realization of this kernel in
BLIS (version 0.9.0) on an 
AMD EPYC 7282 (Rome) processor\footnote{We note that the
realization of \gemm in AMD's native library (AOCL) is basically BLIS in 
disguise; see \url{https://developer.amd.com/amd-aocl/blas-library/}.};
and the tuned implementation of the same kernel in
Intel MKL (version 2021.0) 
on an Intel Xeon Platinum 8358 (Icelake) processor.
In order to generate code specifically tuned for these two architectures, we
only had to set the appropriate target in Part \textcolor{black}{P8} of the TVM
generator. Concretely, lines \textcolor{black}{35--38} in Figure~\ref{lst:basic_GEMM} specify the
backends selected for these two processor architectures. 

\begin{figure}[tbh!]
\begin{center}
        \includegraphics[width=1\textwidth]{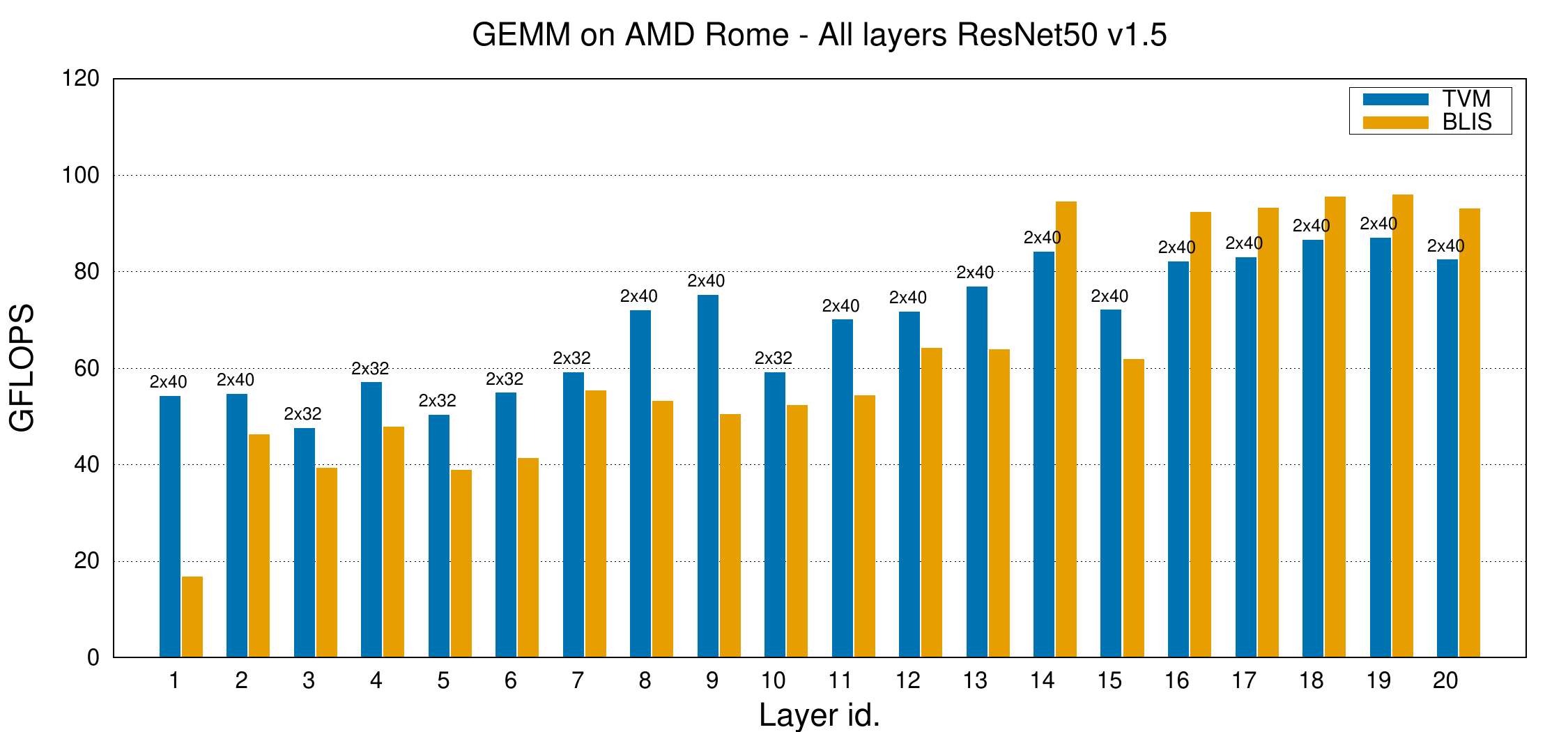}
        \includegraphics[width=1\textwidth]{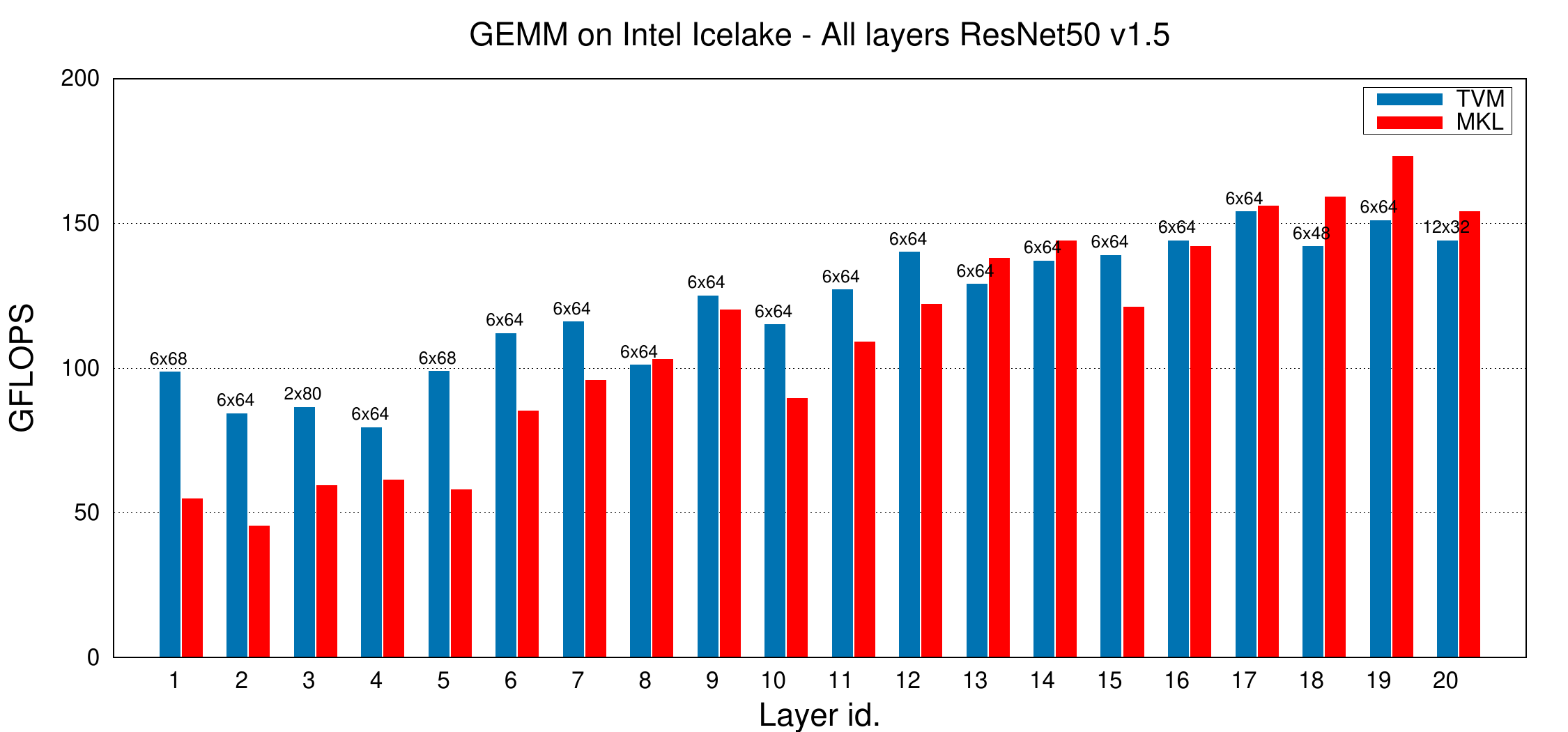}
\end{center}
\caption{Performance of the TVM generator for ResNet-50 v1.5 for AMD and Intel (top and bottom, respectively).}
\label{fig:gemm_amd_intel} 
\end{figure}

Figure~\ref{fig:gemm_amd_intel} 
reports the results for this final experiment using the layers in the ResNet50 model
as testbed. The two charts there show a similar trend, which was
already present in the results for the NVIDIA Carmel core 
in Figure~\ref{fig:microkernel_ARM_ResNet}. Concretely, 
the TVM routine
outperforms the library realizations for the 
``rectangular'' cases,
but it is suboptimal for ``square'' problems.
In order to investigate this behavior in more detail,
given that the library evaluated in the case of the AMD architecture is BLIS, 
we inspected the internals of the micro-kernel integrated in the library 
for that particular processor,
in order to compare it with the micro-kernel generated by TVM. 

Note that the BLIS and TVM solutions
rely on the baseline algorithm for \gemm and, therefore, on a micro-kernel
with \Cresident. Specifically, for the AMD Rome, BLIS hand-encodes a micro-kernel
of dimension $m_r \times n_r = 6\times 16$, unrolling loop 
\texttt{L6} by a factor of 4. 
Given that the AMD Rome features 16 vector registers,
and that the SIMD width is 256 bits (that is, 8 FP32 per vector register) for AVX2,
the BLIS micro-kernel dedicates $6 \times 2 = 12$ vector registers to maintain 
the micro-tile of $C$. Furthermore, at each iteration of loop \texttt{L6}, 
the micro-kernel utilizes  two vector registers to load a row of the micro-panel $B_r$ 
(of dimension $k_c \times 16$ micro-panel) plus 
and single vector register to broadcast one-by-one the six entries of 
a column of the micro-panel $A_r$ (of dimension $6 \times k_c$) prior to operating with
each. In total, the micro-kernel thus occupies 15 out of the 16 available vector registers.
In addition, the code for the micro-kernel features a notable number of
(assembly) prefetching instructions. 

With the above-described configuration for the micro-kernel, the realization
of \gemm in BLIS delivers
95.3 GFLOPS for square \gemm problems of dimension
$m=n=k=$2,000 while, by disabling the prefetching instructions, the 
performance drops to 78.5~GFLOPS.
Compared with that, the best micro-kernel generated with TVM for that problem dimension
corresponds to $m_r \times n_r = 2 \times 40$, which delivers 88.2~GFLOPS.
The flop throughtput rate for TVM is thus somewhere between the two rates
for BLIS (i.e., with and without prefetching) which, on the one hand, 
is not totally surprising as TVM cannot exploit (assembly-level) prefetching instructions.
On the other hand, the dimensions of the best micro-kernel selected by TVM are 
quite surprising. To further investigate this, we instructed TVM to generate a
\gemm routine using a BLIS-like micro-kernel, 
that is, with $m_r \times n_r = 6 \times 16$ and an unrolling factor
of 4. Interestingly, the TVM routine that integrated
this micro-kernel reported 39.60~GFLOPS only.
The reason is that, for this particular micro-kernel dimension and unrolling
factor, TVM produced a micro-kernel that did not maintain the micro-tile of $C$
into the processor registers, incurring into register spilling during the execution
of the micro-kernel loop and considerably degrading performance!
Whether we can enforce TVM to avoid this effect is currently under investigation as,
to a good extent, it depends on  internal behavior of TVM.

\section{Concluding Remarks}
\label{sec:remarks}

We have presented an integral TVM-based solution to automatically obtain 
high performance realizations of \gemm.
On the one hand, our solution departs from conventional library-based realizations of \gemm  in that
the full code is automatically generated, including both the blocked routines for the family of \gemm algorithms extending the ideas of GotoBLAS2 
as well as the processor-specific micro-kernels.
On the other hand, compared with other JIT compilation frameworks, we mimic the techniques
in the GotoBLAS2/BLIS/OpenBLAS2 algorithms for \gemm to obtain blocked algorithms that 
attain an efficient utilization of the cache memories, considerably trimming the
cost of exploring the optimization search space. 
{\color{black} TVM can also generate competitive code for GPUs that perform close to NVIDIA cuBLAS library. However, the schedule for the GPU-specific \gemm differs from that in the CPU version described in this work. Explaining the differences out of scope for this work.}

Our work exposes the programming advantages, 
from the points of view of performance,  maintainability, and portability, 
of the TVM-automatized approach, which can be leveraged, among others, to seamlessly 
generate routines for different data types, 
explore distinct packing schemes, 
evaluate alternative parallelization options, and 
instantiate the entire family of matrix multiplication algorithms.

\section*{Acknowledgments}
This work was supported by the research projects
PID2020-113656RB-C22 (MCIN/AEI/10.13039/ 501100011033), 
and PID2021-126576NB-I00;
and CM via Multiannual
Agreement  with  Complutense  University  in  the  line  Program  to  Stimulate Research for Young Doctors in the context of the V PRICIT under projects PR65/19-22445 and CM S2018/TCS-4423.
A. Castell\'o is a FJC2019-039222-I fellow
supported by MCIN/AEI/10.13039/ 501100011033.
H. Mart\'inez is a postdoctoral fellow supported by the \emph{Consejer\'ia de Transformaci\'on Econ\'omica, Industria, Conocimiento y Universidades de la Junta de Andaluc\'ia}.
This project has received funding from the European High-Performance Computing Joint Undertaking (JU) under grant agreement No
955558 (eFlows4HPC project). The JU receives support from the European Union’s Horizon 2020 research and innovation programme,
and Spain, Germany, France, Italy, Poland, Switzerland, 
and Norway.

\bibliographystyle{ACM-Reference-Format}
\bibliography{article}

\newpage


\end{document}